\documentclass{article}
\usepackage{microtype}
\usepackage{makecell}
\usepackage{graphicx}
\usepackage{subfigure}
\usepackage{booktabs} 

\usepackage{hyperref}
\usepackage{comment}

\newcommand\scalemath[2]{\scalebox{#1}{\mbox{\ensuremath{\displaystyle #2}}}}

\usepackage[accepted]{temp}

\usepackage{amsmath}
\usepackage{amssymb}
\usepackage{mathtools}
\usepackage{amsthm}
\usepackage{comment}
\usepackage{graphicx}   
\usepackage{bm}
\usepackage[flushleft]{threeparttable}
\usepackage{qtree}
\usepackage[free-standing-units=true]{siunitx}
\usepackage[capitalize,noabbrev]{cleveref}

\theoremstyle{plain}
\newtheorem{theorem}{Theorem}[section]
\newtheorem{proposition}[theorem]{Proposition}
\newtheorem{lemma}[theorem]{Lemma}

\theoremstyle{definition}
\newtheorem{definition}[theorem]{Definition}

\theoremstyle{remark}
\newtheorem{remark}[theorem]{Remark}
\usepackage{diagbox}

\usepackage[textsize=tiny]{todonotes}

\icmltitlerunning{Analytical Lyapunov Function Discovery: An RL-based Generative Approach}

\begin{document}

\twocolumn[
\icmltitle{Analytical Lyapunov Function Discovery: An RL-based Generative Approach}



\icmlsetsymbol{equal}{*}

\begin{icmlauthorlist}
\icmlauthor{Haohan Zou}{equal,yyy}
\icmlauthor{Jie Feng}{equal,yyy}
\icmlauthor{Hao Zhao}{zzz}
\icmlauthor{Yuanyuan Shi}{yyy}
\end{icmlauthorlist}

\icmlaffiliation{yyy}{Department of Electrical and Computer Engineering, UC San Diego, La Jolla, United States}
\icmlaffiliation{zzz}{Swiss Federal Institute of Technology in Lausanne}

\icmlcorrespondingauthor{Haohan Zou}{hazou@ucsd.edu}

\icmlkeywords{Machine Learning, ICML}

\vskip 0.3in
]



\printAffiliationsAndNotice{\icmlEqualContribution} 

\begin{abstract}
Despite advances in learning-based methods, finding valid Lyapunov functions for nonlinear dynamical systems remains challenging. Current neural network approaches face two main issues:  challenges in scalable verification and limited interpretability. To address these, we propose an end-to-end framework using transformers to construct analytical Lyapunov functions (local), which simplifies formal verification, enhances interpretability, and provides valuable insights for control engineers. Our framework consists of a transformer-based trainer that generates candidate Lyapunov functions and a falsifier that verifies candidate expressions and refines the model via risk-seeking policy gradient. Unlike \citet{alfarano2024global}, which utilizes pre-training and seeks global Lyapunov functions for low-dimensional systems, our model is trained from scratch via reinforcement learning (RL) and succeeds in finding local Lyapunov functions for \emph{high-dimensional} and \emph{non-polynomial} systems. 
Given the symbolic nature of the Lyapunov function candidates, we employ efficient optimization methods for falsification during training and formal verification tools for the final verification. We demonstrate the efficiency of our approach on a range of nonlinear dynamical systems with up to ten dimensions and show that it can discover Lyapunov functions not previously identified in the control literature. Full implementation is available on \href{https://github.com/JieFeng-cse/Analytical-Lyapunov-Function-Discovery}{Github}.

\end{abstract}

\section{Introduction}
A Lyapunov function is an energy-like function used to certify stability of dynamical systems. A sufficient condition for stability is that the Lyapunov function decreases along system trajectories. Lyapunov functions also play a central role in controller design, providing formal guarantees of closed-loop stability and robustness~\citet{khalil2002nonlinear}.
However, designing a Lyapunov function for nonlinear systems has long been considered more of an `art' than a science, even for stable dynamics, due to its inherent complexities. Motivated by this challenge, we have witnessed great interest in the development of computational algorithms for Lyapunov function construction. \citet{mcgough2010symbolic} employed an evolutionary algorithm for the symbolic computation of Lyapunov functions, but the exponential growth search space impedes its scalability. Alternatively, sum-of-squares (SOS) methods reformulate the task as a semidefinite program (SDP) that certifies stability with polynomial candidates \citep{1470374,papachristodoulou_prajna_2005,sos,dai2023convex}.
However, handling local constraints or non-polynomial dynamics requires auxiliary variables and extra (in)equality constraints \citep{1470374,papachristodoulou_prajna_2005}, which leads to scalability issues of the SOS methods for real-world problems.
Moreover, the theoretical result \citet{ahmadi2011} on asymptotic Lyapunov stability shows that even some very simple globally asymptotically stable dynamics may not agree with a polynomial Lyapunov function of any degree. 

Recent advances in deep learning have enabled data-driven neural Lyapunov function with formal verification \citep{chang2019neural,zhou2022neural, wu2023neural,laypunovsurvey,edwards2024fossil, wang2024lyapunov, yang2024lyapunovstable}. However, these methods face two key challenges: 1) lack of interpretability and 2) high verification costs \citep{laypunovsurvey}. Neural networks' black-box nature limits insights into the system's dynamical behavior. Additionally, over-parameterization and nonlinear activations complicate formal verification, leading to scalability issues. Tools like SMT \citet{chang2019neural}, MIP \citet{wu2023neural}, and $\alpha,\beta$-CROWN \citet{yang2024lyapunovstable} require small, specialized networks to ensure feasible verification times.

\begin{figure}[t]                   
    \includegraphics[width=0.43\textwidth]{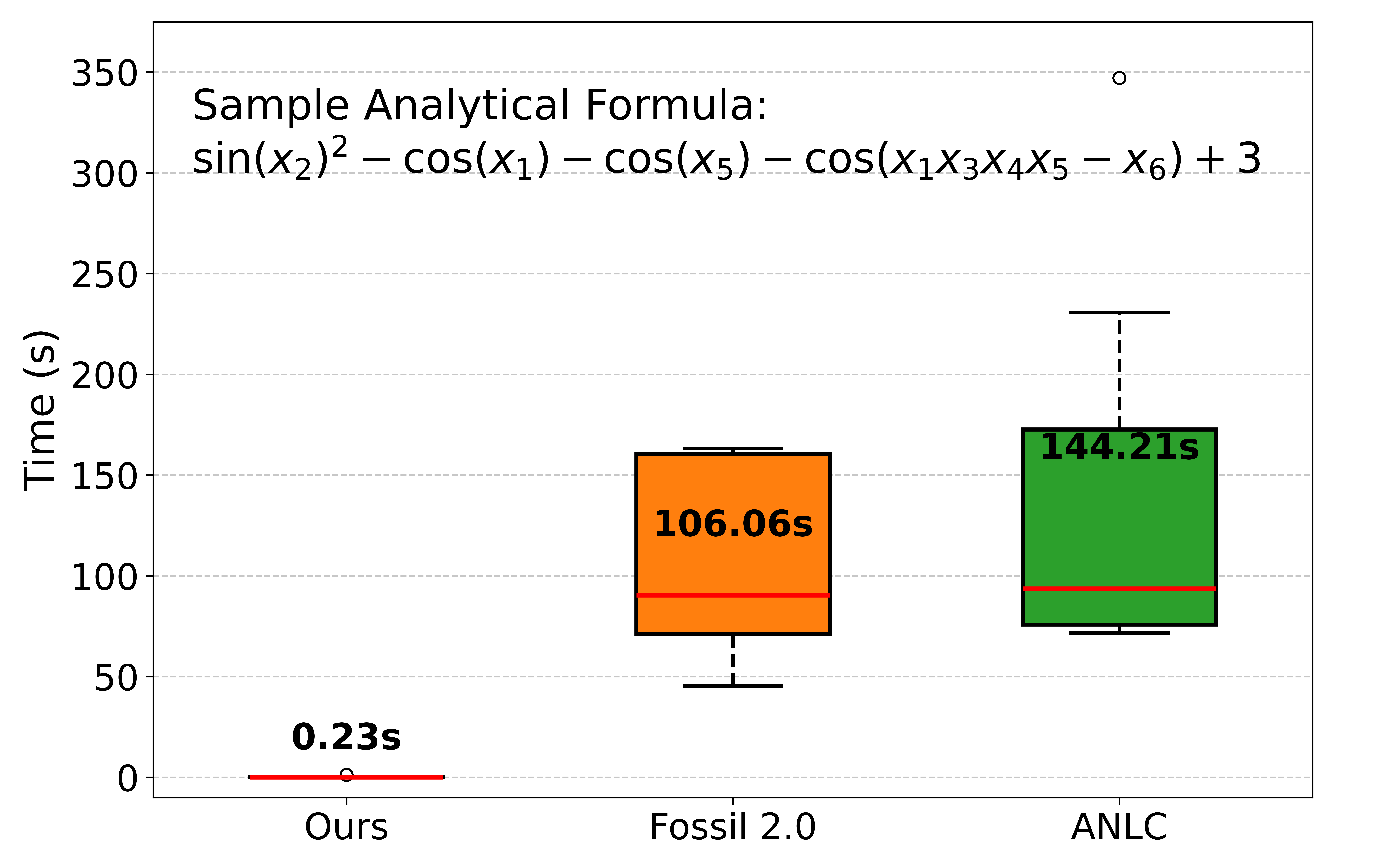}
    \caption{We present the statistics of runtime for single SMT verification (precision $e^{-6}$, tolerance error $e^{-1}$) on the 6-D system (Appendix \ref{subsec: 6d-poly}). Each method is validated once per epoch. The red line represents the median. The mean verification time is averaged over one complete training, and the value is displayed in bold. Neural Network in FOSSIL 2.0 \citet{edwards2024fossil} has a 10-D hidden layer, and Augmented Neural Lyapunov Control (ANLC \citet{ANLC}) has two 10D hidden layers. Our formulas have a complexity (number of tokens) of 20 or fewer. For ours, the best candidate in each epoch is verified in this motivation example.}
    \label{fig: SMT comparison}
\end{figure}

Compared with neural Lyapunov functions, analytical Lyapunov functions offer two distinct advantages. First, their symbolic nature offers interpretability and provides insights for designing stability-guaranteed control policy \citep{sontag1989universal,10163934,10336939,10543148,cui2023structured}. 
Second, analytical functions enable efficient verification due to their simplicity and symbolic structure, which seamlessly integrate with formal verification tools like SMT solvers. This reduces parameter complexity and eliminates the need to verify complex neural network elements like nonlinear activations.
Figure \ref{fig: SMT comparison} highlights the efficiency of verifying analytical expressions compared to neural networks.

In this work, we aim to address the following question:

\emph{Can neural networks effectively discover valid analytical Lyapunov functions directly from complex system dynamics?}

To tackle this challenge, we introduce an end-to-end framework designed to find analytical Lyapunov functions for nonlinear dynamical systems given in analytical form.
Building on the transformer’s ability to model complex dependencies \citet{transformer} and the success of deep symbolic regression methods \citet{dgsr},
our framework deploys a symbolic transformer \citet{E2E} for Lyapunov function discovery.
The transformer’s encoder captures system dynamics represented as token sequences derived from the ordinary differential equations (ODEs), while the decoder generates candidate Lyapunov functions by modeling symbolic token distributions. Given the lack of high-quality (local) Lyapunov function datasets, particularly for high-dimensional systems, we propose a reinforcement learning (RL)-based approach to search for Lyapunov functions on a per-system basis, instead of pre-training like \citet{alfarano2024global}. 
We verify Lyapunov conditions by localized sampling in the neighborhoods of minimizers of the candidate expressions, which are most likely to have violations. The identified counterexamples are then incorporated into the training set for further optimization. Our main contributions are summarized as follows:
\begin{itemize}
    \item We introduce the first RL-based framework for directly discovering analytical Lyapunov functions for nonlinear dynamical systems, bypassing the need for supervised learning with large-scale datasets.  
    \item We propose a novel and efficient policy optimization pipeline integrating global-optimization-based Lyapunov verification, reward design for candidate Lyapunov evaluation, and risk-seeking policy gradient to optimize the symbolic transformer, 
    trainable on machines with limited computation resources. 
    \item We demonstrate the efficiency of our method on various systems, including non-polynomial dynamics like the pendulum, quadrotor, and power system frequency control. Notably, our approach scales to a 10-D system and discovers a valid local Lyapunov function for power system frequency control with lossy transmission lines, which is previously unknown in the literature. 
\end{itemize}

\section{Related Work}
\label{subsec:Lyapunov_function_related_work}
\subsection{Learning-based Lyapunov Function Construction}

The field of learning-based Lyapunov function construction is advancing rapidly. \citet{chang2019neural} formulated Lyapunov condition violations as the objective, jointly learning a neural Lyapunov function and a linear controller to guarantee stability for a given system, with stability verified via SMT solvers. \citet{zhou2022neural} extended this to unknown dynamics with a neural controller. \citet{dailyapunov} and \citet{wu2023neural} focused on discrete-time systems, using MIP solvers for stability verification, requiring piecewise linear approximations. \citet{yang2024lyapunovstable} applied $\alpha,\beta$-CROWN for scalable neural network verification, extending state feedback to output feedback control. However, scalability remains a challenge: SMT solvers handle up to 30 neurons, MIP solvers scale to 200 neurons \citet{DawsonSurvey}, and $\alpha,\beta$-CROWN \citet{yang2024lyapunovstable} is limited to a three-layer architecture (16 neurons per layer).

In contrast to neural Lyapunov functions, \citet{feng2024} and \citet{alfarano2024global} derived analytical Lyapunov functions. \citet{feng2024} combined a neural network with the symbolic regression package \textit{PySR} \citep{pysr}, which approximates the network to produce analytical Lyapunov functions, but the lack of interaction between system dynamics and symbolic regression component limits its potential. \citet{alfarano2024global} pre-trained a transformer on backward- and forward-generated global Lyapunov function datasets, relying on beam search for candidate generation. However, their method cannot adaptively refine the candidate Lyapunov functions if the beam search fails on specific dynamics, and it requires a dataset that is expensive to generate (e.g., thousands of CPU hours for a 5-D dynamics dataset) to achieve adequate generalization during inference. Furthermore, its emphasis on global stability limits its applicability to real-world, nonpolynomial control systems, which typically only admit local stability. Consequently, an RL-based approach that directly searches Lyapunov functions for a given system is indispensable.

\subsection{Symbolic Regression with Generative Model}
\label{subsec:sr-work}
Symbolic regression is a supervised learning task, searching for an analytical function $f: \mathbb{R}^n \to \mathbb{R}$ that fits $y_i \in \mathbb{R}$ from input $x_i \in \mathbb{R}^n$ \citep{dso}. With suitable extensions, it can scale to multi-input–multi-output mappings.

\textbf{RL-based Symbolic Regression.} 
RL-based symbolic regression algorithms 
\citet{dso, 2020costa, dso21} employed generative models, typically RNNs, to generate distributions of symbolic tokens representing mathematical operations and variables, from which analytical expressions are sampled. Rewards, evaluating the quality of the sampled expressions, are measured by fitness metrics like RMSE. Due to the non-differentiable step of converting token sequences into symbolic equations, policy gradients are used to optimize the output distributions. \citet{rlgp2021} extended this approach by integrating Genetic Programming (GP) to refine generated expressions, improving the overall performance.

\textbf{Pre-trained Symbolic Regression methods.} 
These methods are inspired by the success of transformers in Natural Language Processing (NLP) tasks. These algorithms contain two steps: 1) pre-train an encoder-decoder network to model $p(f|\mathcal{D})$ on curated datasets by cross-entropy loss, and 2) sample from this distribution during inference via beam search or Monte Carlo Tree Search (MCTS). Methods like \citet{NSR, E2E, NSRwH} rely on beam search without gradient refinement, often yielding suboptimal results for out-of-distribution data. In contrast, \citet{dgsr} integrates RL-based policy gradient optimization with end-to-end RMSE loss for both pre-training and inference, allowing gradient refinement for unseen datasets during inference. Further, \citet{tpsr, dgsr-mcts} enhance the decoding process (expression generation) by incorporating MCTS with feedback, such as fitting accuracy and equation complexity.

\section{Preliminary}
Our framework searches for analytical Lyapunov functions for autonomous nonlinear dynamical systems at an equilibrium point. Without loss of generality, we assume the origin to be the equilibrium point. 

\begin{definition}[Dynamical systems]
\label{def:dynamical system}
An $n$-dimensional autonomous nonlinear dynamical system is formulated as 
\begin{equation}\label{eq:nonlinear_dynamics}
    \frac{dx}{dt}=f(x)\,,x(0)=x_0,
\end{equation}
where $f:\mathcal{D}\to \mathbb{R}^n$ is a Lipschitz-continuous vector field, and $\mathcal{D}\subseteq \mathbb{R}^n$ is a set containing the origin that defines the state space. Each $x(t)\in\mathcal{D}$ is a state vector. 
\end{definition}

\begin{definition}[Asymptotic stability]
    \label{def:asm_stb}
    A system of \eqref{eq:nonlinear_dynamics} is stable at the origin if $\forall \:\epsilon>0$, there exists $\delta = \delta(\epsilon)>0$ such that $\lVert x(0)\rVert<\delta \implies \lVert x(t)\rVert<\epsilon$, $\forall \: t\geq 0$. The origin is asymptotically stable if it is stable and $\delta$ can be chosen such that $\lVert x(0)\rVert <\delta \implies \lim\limits_{t \to \infty}x(t)=0$ \citep{khalil2002nonlinear}.
\end{definition}

\begin{definition}[Lie derivative] \label{def:lie}
The Lie derivative of a continuously differentiable scalar function $V:\mathcal{D}\to\mathbb{R}$ along the trajectory of \eqref{eq:nonlinear_dynamics} is given by
\begin{equation}
    \label{eq:Lie}
    L_f V(x)=\sum_{i=1}^n\frac{\partial V}{\partial x_i}\frac{dx_i}{dt}=\sum_{i=1}^n\frac{\partial V}{\partial x_i}f_i(x).
\end{equation}
\end{definition}

\begin{proposition}[Lyapunov functions for asymptotic stability]\label{pro:lyap} Let $x = 0$ be an equilibrium point for \eqref{eq:nonlinear_dynamics} and $\mathcal{D}\subseteq\mathbb{R}^n$ be a domain containing the $x = 0$. Let $V:\mathcal{D}\to\mathbb{R}$ be a continuously differentiable function such that
\begin{subequations}
    \begin{align}
        &V(0)=0 \text{ and } V(x)>0 \text{ in } D\backslash\{0\}\label{eq:Lyapunov},\\ 
        &L_f V(x)< 0\text{ in } D\backslash \{0\},\label{eq:decrease}
    \end{align}
\end{subequations}
then the origin is asymptotically stable.
\end{proposition}

\begin{definition}[Lyapunov risk]
\label{def:risk}
Consider a candidate Lyapunov function $\Tilde{V}$ for system $f$ from Definition~\ref{def:dynamical system}. For a dataset $\mathcal{X} = \{x_1, \cdots, x_N\}$ 
where $x_i \in \mathcal{D}$, the Lyapunov risk of $\Tilde{V}$ \citet{chang2019neural} over $\mathcal{D}$ is defined by 
\begin{equation}
\label{eq:risk} 
     \scalemath{0.9}{\mathcal{L}(\Tilde{V}) = \frac{1}{N}\sum_{i=1}^N\left (\max(0, L_{f} \Tilde{V}(x_i)) + \max(0, -\Tilde{V}(x_i)) \right).}
\end{equation}
\end{definition}

\section{Proposed Framework}
\label{sec:framework}

\begin{figure*}[h!]
    \centering
    \includegraphics[width=0.9\textwidth]{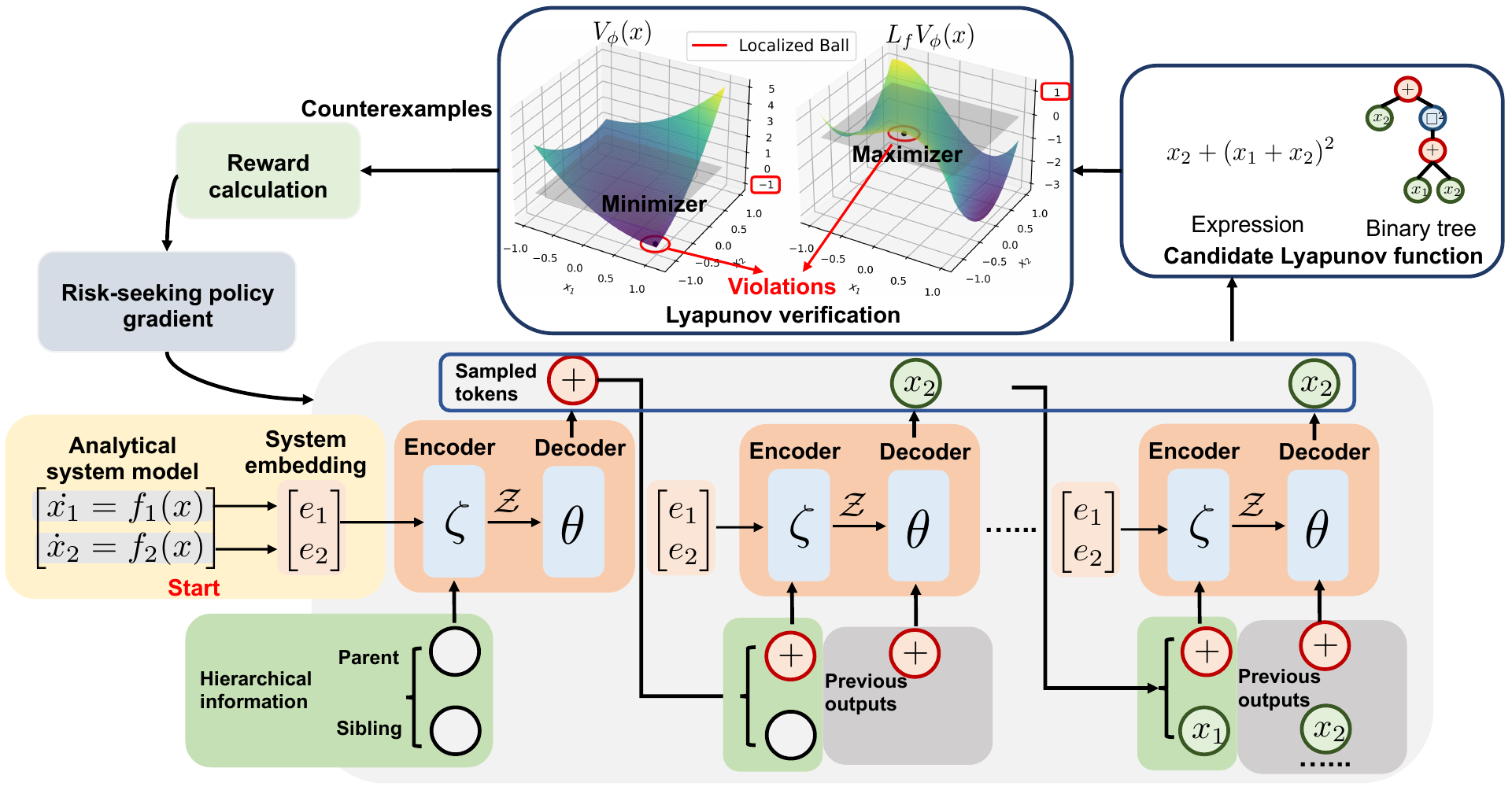}
    \caption{ \textbf{Framework overview}: The transformer takes embeddings of the dynamical system model as input and generates candidate Lyapunov functions in an autoregressive manner. Hierarchical information is deployed to enhance the model input. For example, when generating the last token $x_2$, its parent is $+$, and its sibling is $x_1$. The output is the pre-order traversal of the expression's binary tree. Candidates are verified using a global-optimization-based verification process, with counterexamples added to the training set for reward calculation. The transformer is updated via the risk-seeking policy gradient. The program terminates once a valid expression is found.}
    \label{fig:framework}
\end{figure*}

This section introduces our RL-based generative approach, which aims to find an analytical Lyapunov function for a given dynamical system that certifies asymptotic stability following conditions in Proposition \ref{pro:lyap}.  Successfully identifying such a function guarantees system stability.

The framework, visualized in Figure \ref{fig:framework}, consists of three components: 1) a symbolic transformer, parameterized as $\phi=\{\zeta, \theta\}$, for candidate analytical Lyapunov functions generation, where $\zeta$ and $\theta$ denote the parameters of encoder and decoder, 2) a numerical verifier employing the SHGO \citet{shgo} global optimization algorithm for Lyapunov conditions' checking (Proposition \ref{pro:lyap}) and counterexamples' feedback, and 3) a risk-seeking policy gradient algorithm optimizing the transformer's parameters based on candidate Lyapunov functions' rewards. To tackle the challenges posed by the exponentially growing search space of complex, high-dimensional systems, our framework integrates Genetic Programming as expert guidance to improve expression quality and training efficiency. We denote $\mathcal{X} \subseteq \mathcal{D}$ as the training set for reward calculation.
\subsection{Candidate Lyapunov Function Generation from Symbolic Transformer}
\label{subsec: transformer}

\textbf{Expression Representation.} Inspired by the deep symbolic regression frameworks, we use a symbolic transformer model as the backbone. The transformer takes a dynamical system $f(x)$ as input and generates candidate analytical Lyapunov functions $\Tilde{V}_{\phi}$ such that: $\Tilde{V}_{\phi}(x)>0 \text{ and } L_{f}\Tilde{V}_{\phi} < 0, \forall \: x \in \mathcal{D}\backslash\{0\}$. Following the expression representation rules in \citet{Lample2020Deep}, our framework represents symbolic transformer models' input and output as sequences of symbolic tokens. Each mathematical expression can be converted into a symbolic expression tree, a binary tree where internal nodes are symbolic operators and terminal nodes (leaves in the tree) are variables or constants. Symbolic operators can be either unary (i.e., one child), such as $\sin, \cos$, or binary (i.e., two children), such as $+, \times$. Furthermore, each symbolic expression tree can be represented as a sequence of node values, either symbolic tokens or numerical coefficients, by its pre-order traversal (i.e., first visiting the parent, then traversing the left child and right child). In this way, each expression obtains a pre-order traversal representation, which can uniquely reconstruct the original expression \citet{dso}. 
We denote $\Tilde{V}_{\phi_{i}}$ as the $i^\textrm{\tiny th}$ node value in the pre-order traversal of $\Tilde{V}_{\phi}$'s expression tree, and $\mathcal{L}_s$ as the symbolic library, e.g. $\{+, \times, \log, \sin, x_j\}$, where $\Tilde{V}_{\phi_{i}}$ is sampled from.

\textbf{Dynamics Tokenization.} By Definition \ref{def:dynamical system}, the symbolic transformer models' input $f(x)$ is composed by $n$ ordinary differential equations $\frac{dx_i}{dt} = f_i(x)$ for $i = 1, \cdots, n$. Each analytical expression $f_i(x)$ can be represented as a sequence of symbolic tokens and numerical coefficients by the pre-order traversal of its expression tree. 
Concatenated the sequences of pre-order traversal for all $f_i(x)$, with \texttt{SOS} (start token) and \texttt{EOS} (end token) as separators, we obtain the tokenized dynamics, which is fed into the encoder of symbolic transformer and encoded as a latent vector $\mathcal{F} \in \mathbb{R}^{p}$.
 The numerical coefficients are tokenized in two schemes: an integer is represented as a sequence of digits in base $b=10$ (e.g. 123 is tokenized as $[1, 2, 3]$), and a real number is represented in scientific notation rounded to 4 significant digits (e.g. 3.14 is tokenized as $[3, 1, 4, 0, 10^{0}]$). A detailed example is illustrated in Figure \ref{fig:pendulum_example}, Appendix \ref{app: transformer}, where we present the symbolic representations of the simple pendulum dynamics in sequences of pre-order traversal.

\textbf{Candidate Expression Generation.} The decoder generates candidate expressions $\Tilde{V}_{\phi}$ in an auto-regressive manner. That is, each token $\Tilde{V}_{\phi_{i}}$ in the pre-order traversal of $\Tilde{V}_{\phi}$ is sampled from the symbolic library $\mathcal{L}_s$ according to conditional distribution $p(\Tilde{V}_{\phi_{i}}|\Tilde{V}_{\phi_{1:(i-1)}},\phi, f(x))$, where $\Tilde{V}_{\phi_{1:(i-1)}}$ is the first $(i-1)$ selected symbolic tokens in the pre-order traversal of $\Tilde{V}_{\phi}$.  This conditional dependence can be achieved by the decoder, which emits a probability distribution $\psi$ over the symbolic library $\mathcal{L}_s$, conditioned on the previously selected tokens and input dynamics. Since analytical expression in its pre-order traversal is inherently hierarchical, we also deploy the hierarchical tree state representation \citep{dso, dgsr}. This method enhances the decoder input by concatenating the representations of the parent and sibling nodes with previously selected outputs and the dynamics.
Once the token sampling process for $\Tilde{V}_{\phi}$ is complete, we evaluate the function value at origin $\Tilde{V}_{\phi}(0)$ and subtract it from $\Tilde{V}_{\phi}$, ensuring the Lyapunov condition $\Tilde{V}_{\phi} (0) = 0$. Through this process, we can sample a batch of $Q$ candidates 
$\Tilde{\mathcal{V}}_{\phi}=\{\Tilde{V}_{\phi}^{i} \sim p(\Tilde{V}_{\phi} |\phi, f(x))\}_{i=1}^Q$, which will be verified according to the Lyapunov conditions.

\subsection{Verification and Falsification Feedback}
\label{sub:ver} 
Leveraging the analytical nature of candidate Lyapunov functions, efficient methods for symbolic expressions, such as root finding \citet{feng2024}, can be applied to Lyapunov condition verification and counterexample generation. In this work, we propose a global-optimization-based numerical verification algorithm, using Simplicial Homology Global Optimization (SHGO) \citep{shgo}, that effectively checks the Lyapunov conditions around minimizers and feedback counterexamples into the training set $\mathcal{X}$ for reward calculation. SHGO is a constrained global optimization algorithm with theoretical guarantees for convergence. To make the paper self-contained, we present the guarantees in Proposition \ref{pro:shgo}. This algorithm identifies the global minimizer over the state space from a set of local minimums, each of which is obtained from a convex sub-domain in the feasible search space. Taking advantage of the theoretical results, we employ the SHGO algorithm for counterexample detection in our verification process.

\begin{proposition}[Convergence Gurantees of SHGO \citet{shgo}]\label{pro:shgo} 
For a given continuous objective function $f$ that is adequately sampled by a sampling set of size $N_s$. If the size of the minimizer pool $\mathcal{M}$ extracted from the directed simplex (a convex polyhedron) $\mathcal{H}$ is $|\mathcal{M}|$. Then any further increase of the sampling size $N_s$ will not increase $|\mathcal{M}|$.
\end{proposition} 

This result shows that if the initial points are adequately sampled, that is, the union of identified locally sub-convex domains initiated from starting points covers the feasible search space, then the minimizer pool $\mathcal{M}$, which contains all local minimum extracted from the directed simplexes, will contain the global minimizer of the feasible search space. Notably, the required sampling size $N_s$ can be unbounded.

During verification, for a candidate $\Tilde{V}_{\phi}$, SHGO is first applied to identify minimizers $x_{1}^{*}$ and $x_{2}^{*}$ of $\Tilde{V}_{\phi}$ and its negated Lie derivative $-L_{f}\Tilde{V}_{\phi}$ in the state space $\mathcal{D}$. These minimizers highlight the regions where Lyapunov conditions are most likely to fail. Next, data points $x$ from neighborhoods around these minimizers, $\mathcal{B}_{r}(x_{1}^{*})$ and $\mathcal{B}_{r}(x_{2}^{*})$ where $r$ is a small value relative to $ \|\mathcal{D}||_{2}$, are sampled to check Lyapunov conditions, i.e. $\Tilde{V}_{\phi}(x) > 0$ and $- L_{f}\Tilde{V}_{\phi}(x) > 0$ for $x \in \mathcal{D}\backslash \{0\}$. This localized sampling scheme effectively identifies violations within $\mathcal{D}$. Additional random sampling covering approximately 30\% of the total data and condition checking across the state space are performed to complement this localized sampling to capture additional potential violations and provide a global assessment. Identified counterexamples $\mathcal{X}_{ce}$ are added to the training set $\mathcal{X}$ for reward calculations. Once a candidate Lyapunov function passes this verification process and does not encounter any violation in $\mathcal{X}$, it indicates a numerically valid solution is found, pending the final formal verification. Appendix \ref{app:verification} details the verification implementation. 

\subsection{Risk-Seeking Policy Gradient}\label{subsec:risk-seeking}
The empirical Lyapunov risk $\mathcal{L}(\Tilde{V}_{\phi})$ in Equation \eqref{eq:risk} quantifies the violation degree of Lyapunov conditions over a given dataset. Following \citet{dso, bastiani2024complexityaware}, we apply the continuous mapping $g(x) = \frac{1}{1+x}$ and define proposed Lyapunov risk reward as: 
\begin{align}
\label{eq: reward function}
    R(\Tilde{V}_{\phi}) &= g\left(\mathcal{L}(\Tilde{V}_{\phi})\right) 
    =\frac{1}{1 + \mathcal{L}(\Tilde{V}_{\phi})},
\end{align}
where $\mathcal{L}(\Tilde{V}_{\phi})$ is measured over training set $\mathcal{X}$. The continuous mapping $g(x)$ bounds the reward value to $[0,1]$. 
For candidate expressions that do not incorporate all state variables or are analytically incomplete, we assign their rewards to be $0$ to ensure they are effectively penalized.

Given the violation measure for each sampled $\Tilde{V}_{\phi}$ is non-differentiable with respect to the transformer parameters $\phi$, we employ the risk-seeking policy gradient to update $\phi$ end-to-end. The objective of standard policy gradient \citet{REINFORCE} is defined to maximize $J_{\text{std}}(\phi) = \mathbb{E}_{\Tilde{V}_{\phi} \sim p(\Tilde{V}_{\phi} | \phi, f(x))} [ R(\Tilde{V}_{\phi})]$, the expectation of the reward function $R(\cdot)$ for candidates' quality evaluation based on the current parameter $\phi$. This objective is desirable for problems aiming to optimize the average performance of the policy network. In our task, final performance depends on finding \emph{at least} one valid Lyapunov function that meets the conditions in Proposition~\ref{pro:lyap}, rather than optimizing for average performance. Consequently, standard policy gradient methods are inadequate due to the misalignment.

In our framework, we adopt risk-seeking policy gradient \citet{dso} that only focuses on maximizing best-case performance. Let $R_{\alpha} (\phi)$ as the $1 - \alpha$ quantile of the distribution of rewards of sampled candidates under the current policy $\phi$. The learning objective of risk-seeking policy gradient, parameterized by $\alpha$, is formulated as:
\begin{equation}
\label{eq: risk pg}
    \scalemath{0.89}{J_{\text{risk}}(\phi, \alpha) =  \mathbb{E}_{\Tilde{V}_{\phi} \sim p(\Tilde{V}_{\phi} | \phi,f(x))}\left[ -R(\Tilde{V}_{\phi}) \mid R(\Tilde{V}_{\phi}) \geq R_{\alpha} (\phi) \right]}.  \\
\end{equation}
This objective aims to optimize only the rewards of high-quality candidates from the top $1 - \alpha$ quantile.

\begin{algorithm}[t!]
    \caption{Training Framework for Analytical Lyapunov Function Discovery via Reinforcement Learning}\label{alg:framework}
    \textbf{Input:}  Dynamics $f(x)$, state space $\mathcal{D}$, quantile $\alpha$, symbolic library $\mathcal{L}$, batch size $Q$, max complexity $k$, radius $r$. \\
    \textbf{Output:} Valid Lyapunov function $\mathcal{V}^{*}$ for system $f(x)$. 
    
    \begin{algorithmic}[1]
        \STATE Initialize the conditional generator with parameters $\phi$,
        \STATE Randomly sample training datapoints $\mathcal{X} = \{x_1, \cdots, x_N \} \text{ where }x_i\in\mathcal{D}$, 
        \WHILE{no valid candidates found}
        \STATE $ \Tilde{\mathcal{V}}_{\phi} \leftarrow \{\Tilde{V}_{\phi}^{i} \sim p(\Tilde{V}_{\phi} |\phi, f(x))\}_{i=1}^Q$, 
       \STATE $\Tilde{\mathcal{V}}_{gp} \leftarrow \text{Genetic Programming}(\Tilde{\mathcal{V}}_{\phi})$,
        \STATE $\Tilde{\mathcal{V}} \gets \Tilde{\mathcal{V}}_{\phi} \; \cup \;\Tilde{\mathcal{V}}_{gp}$,
        \STATE $\mathcal{V}^{*}, \mathcal{X}_{ce} \leftarrow 
        \text{verification} (\mathcal{\Tilde{V}}, r, \mathcal{D})$,
        \COMMENT{Verify candidates $\mathcal{\Tilde{V}}$, return the valid Lyapunov function $\mathcal{V}^{*}$ (if any) and counterexamples $\mathcal{X}_{ce}$. Details in App. \ref{app:verification}. } 
        \IF{$\mathcal{V}^{*}$ is not empty}
            \STATE \textbf{Return} $\mathcal{V}^{*}$. 
        \ENDIF
        \STATE $\mathcal{R} \gets \{R(\Tilde{V}^{i}) \; \forall \; \Tilde{V}^{i} \in \Tilde{\mathcal{V}} \}$,
        \STATE $R_\alpha(\phi) \leftarrow (1 - \alpha)$-quantile of $\mathcal{R}$,
        \STATE $\phi \gets \phi - \nabla_\phi J_{\text{risk}}(\phi, \alpha)$ ,
        \COMMENT{risk-seeking policy gradient update. Equation \eqref{eq: risk pg} }
        \STATE $\phi \gets \phi - \nabla_\phi \mathcal{L}(\Tilde{\mathcal{V}}_{gp})$ ,
        \COMMENT{expert guidance loss based on the genetic programming refined Lyapunov functions $\Tilde{\mathcal{V}}_{gp}$ to update policy. Equation \eqref{eq: weighted ce}.}
        \STATE Concatenate counterexamples $\mathcal{X}_{ce}$ to dataset $\mathcal{X}$.
    \ENDWHILE
    \end{algorithmic}
\end{algorithm}

\subsection{Automated Expert Guidance}
\label{subsec: gp}
While the risk-seeking policy gradient algorithm effectively optimizes the model, training efficiency can be enhanced by off-the-shelf tools that further explore the function space based on the transformers. Inspired by \citet{rlgp2021}, we incorporate a Genetic Programming (GP) component (DEAP \citet{fortin2012deap}) into the training paradigm.

The GP algorithm starts with a batch of initial populations (expression trees) and iteratively refines these populations through evolutionary operations: mutation, selection, and crossover, with a pre-defined metric to evaluate the fitness of populations to the task (e.g., MSE for symbolic regression task). Within our framework, we feed the latest batch of generated candidates $\{\Tilde{V}_{\phi}^{i} \sim p(\Tilde{V}_{\phi} |\phi, f(x))\}_{i=1}^Q$ from the decoder into the GP module as the starting population\footnote{Without a good initial population from the transformer, GP algorithms itself face significant challenges in directly finding valid Lyapunov functions for high-dimensional systems due to the exponentially growing search space. Appendix \ref{app: ablation_gp}.}, refine these expressions through evolutionary operations with Lyapunov risk reward as the measure of fitness, and obtain a batch of refined expressions. We select an `elite set' of the refined expressions $\Tilde{\mathcal{V}}_{gp} = \{\Tilde{V}_{gp}^{i} \sim \text{GP}(\Tilde{V}_{\phi})\}_{i = 1}^{G}$, regard them as target expressions,
and optimize the transformer model in a supervised learning manner, 
maximizing the probability that the generated token matches the reference tokens from $\Tilde{V}_{gp}$, with the following expert guidance loss:
\begin{equation}
\label{eq: weighted ce}
    \scalemath{0.77}{\mathcal{L}(\Tilde{\mathcal{V}}_{gp}) = \frac{1}{G} \sum_{i=1}^{G} \frac{1}{k_i} R(\Tilde{V}_{gp}^{i}) \sum_{j = 1}
    ^{k_i} -\log \left(p(\Tilde{V}_{gp_j}^{i}|\Tilde{V}_{gp_{1:(j-1)}}^{i},\phi,f(x))\right),}
\end{equation}
where $G$ is the number of expressions in $\Tilde{\mathcal{V}}_{gp}$, and $k_i$ is the complexity (number of symbolic tokens in the pre-order traversal) of $\Tilde{V}_{gp}^{i}$. Each expression $\Tilde{V}_{gp}^{i}$ is weighted by its Lyapunov risk reward in $\mathcal{L}$. Algorithm \ref{alg:framework} summarizes the training process, with more details in Appendix \ref{app:gp}. 
The GP solutions explore the characteristics of Lyapunov functions that have not been captured by the transformer yet and effectively guide the transformer. \begin{remark}[Exploration–exploitation trade-off]
    In our framework, exploration arises during candidate expression generation and through GP's evolutionary operations, while exploitation is driven by the risk-seeking policy gradient and the expert-guided loss.
\end{remark}

\section{Experiment}
\label{sec:exp}

We validate the proposed algorithm across a variety of nonlinear dynamics by finding their local Lyapunov functions at the equilibrium point to verify their stability, where the systems are autonomous (or closed-loop systems with known feedback control laws).
We use \textit{dReal} \citet{dReal} SMT solver for final verification of found Lyapunov functions, with a numerical tolerance error $\epsilon = e^{-3}$ and precision $\delta = e^{-12}$, over the state space, i.e. $V(x) > \delta$ and $L_{f}V(x) < - \delta$ over  $\mathcal{D} \backslash \mathcal{B}_{\epsilon}(0)$. The excluded ball $\mathcal{B}_{\epsilon}(0)$ is to avoid numerical issues, which is a common practice for SMT-based formal verification \citep{chang2019neural}. Global stability can be determined through further expert analysis; for instance, if the Lie derivative is a negation of SOS, it is sufficient to establish global stability.

\subsection{Experimental Setting}

\begin{table*}[h]
\caption{Performance and time consumption of our method on test dynamics. `App.' refers to Appendix.}
\vskip 0.15in
\centering 
\begin{tabular}{@{}llllll@{}}
\toprule
                      \textbf{Dynamics}& \textbf{Runtime}&\textbf{Ver.$^a$}&  \textbf{Found Lyapunov Functions} & \textbf{Stab}$^\ddagger$
                      &\textbf{Succ \% $^\ddagger$}\\ \midrule

\multicolumn{1}{l|}{2-D Polynomial Sys {\tiny(App. \ref{app:meta_1})}} & 68s & $2$ ms & $V = 9 x_1^2 + 2x_2^2$ & l.a.s. & 100 \\

\multicolumn{1}{l|}{2-D Van Der Pol {\tiny(App. \ref{app:van_der_pol})}} &   126s & $1$ ms & $V=x_1^2 + x_{2}^2$ & l.a.s. & 100 \\ 

\multicolumn{1}{l|}{2-D Simple Pendulum {\tiny(App. \ref{app:pendulum})}} & 288s & $1$ ms & $V = 2(1-\cos(x_1)) + x_2^2$ & l.a.s. & 100 \\
\midrule
\multicolumn{1}{l|}{3-D Polynomial Sys {\tiny(App. \ref{app:meta_2})}}  &   112s & $1$ ms &  $V = 9x_1^2 + x_2^2 + x_3^2$ & l.a.s. & 100\\ 

\multicolumn{1}{l|}{3-D Trig Dynamics {\tiny(App. \ref{app: 3-D Trig Dynamics})}}  &   157s & $1$ ms &  $V = 1 - \cos(x_1)^2 + x_2^2 + \sin(x_3)^2$ & l.a.s. & 100\\ 
\midrule
\multicolumn{1}{l|}{4-D Lossy Power Sys {\tiny(App. \ref{app:lossy_power})}}  &   3632s & $621$s &  $V = \omega_1^2 + \omega_2^2 + (\omega_2 - \sin(\delta_1) + \sin(\delta_2))^2$ & l.a.s. & 100\\  
\midrule
\multicolumn{1}{l|}{6-D Polynomial Sys {\tiny(App. \ref{subsec: 6d-poly})}}  & 1667s & $7$ ms & $V = \sum_{i=1}^{6} x_i^2$ & l.a.s. & 100\\ 
\multicolumn{1}{l|}{6-D Quadrotor {\tiny(App. \ref{app:quadrotor})}}  & 3218s & $1$ ms & $V = \sum_{i=1}^{6} x_i^2$ & g.a.s. & 80\\ 
\multicolumn{1}{l|}{6-D Lossless Power Sys {\tiny(App. \ref{app:lossless_power})}}  & 18094s & $2$ ms & $\begin{array}[t]{l} V = (\sum_{i=1}^{3} \omega_i^2) - \\0.5\left(\sum\limits_{i=1}^{3} \sum\limits_{\substack{j=1, i\neq j}}^{3} \cos(\delta_i - \delta_j) - 1 \right) \end{array}$ & l.a.s. & 60\\ 
\midrule
\multicolumn{1}{l|}{9-D Synthetic Sys \tiny{(App. \ref{app:9-d_synthetic})}}  & 27047s & $6.6$ s & $\begin{array}[t]{l} V = \left(\sum_{i=1}^{6} x_i^2\right) + \sin(x_7)^2\\ + x_8^2 -\cos(x_9)+1\end{array}$ & l.a.s. & 60\\ 
\midrule
\multicolumn{1}{l|}{10-D Polynomial Sys {\tiny (App. \ref{app:10d})}}  & 64223s & $2$ ms & $V = \sum_{i=1}^{10} x_i^2$ & l.a.s. & 60\\ 
                 
\bottomrule
    \multicolumn{5}{l}{\tiny $a$. `Ver.' presents the time consumption for the final verification of the found Lyapunov functions. All found Lyapunov functions passed SMT solver's verification.} \\ 
    
    \multicolumn{5}{l}{\tiny  $\ddagger$. `Stab' means stability. In this column, `g.a.s' represents globally asymptotically stable, and `l.a.s.' represents locally asymptotically stable.}\\
    
    \multicolumn{5}{l}{\tiny $\ddagger$. 'Succ \%' denotes the successful rate of finding a valid Lyapunov function out of 5 random seeds. } \\

\end{tabular}
\label{tab:performance summary}
\end{table*}

\textbf{Test Dynamics.}
We categorize our collected test dynamical systems into two kinds: 1). Polynomial Systems, 
and  2). Non-polynomial Systems, where three polynomial systems are adopted from \citet{alfarano2024global} (Appendices \ref{app:meta_1} \& \ref{app:meta_2}), and others come from real-world examples. Detailed information about systems is summarized in Appendices \ref{poly} and \ref{nonpoly}. The dimension of these test systems ranges up to 10.

\textbf{Implementation Details of Framework.}
Detailed explanation for all components in our framework is presented in Appendices \ref{app: transformer} (Transformer), \ref{app:verification} (Global-optimization-based Numerical Verification), \ref{app:pg} (Risk-seeking Policy Gradient), and \ref{app:gp} (Genetic Programming).
The symbolic library $\mathcal{L}_s$ is defined as $\{+, -, \times, \sin, \cos, x_i \}$ in all tests.

\textbf{Baseline Algorithms.}
We compare our proposed framework against four baseline algorithms for continuous nonlinear dynamics. \textit{Neural methods}: 1) Augmented Neural Lyapunov Control (ANLC) \citet{ANLC}, and 2) FOSSIL 2.0 \citet{edwards2024fossil}. \textit{Analytical methods}: 3) the transformer-based global Lyapunov search of \citet{alfarano2024global}, and 4) SOS methods via SOSTOOLS (Matlab) \citep{sostools}. We use the formulation of \citet{1470374} on polynomial systems, and apply the recasting technique of \citet{papachristodoulou_prajna_2005} to convert non-polynomial dynamics into rational form so they can also be handled by SOS.

Both ANLC and FOSSIL 2.0 train a neural Lyapunov function with the empirical Lyapunov risk loss and employ a counter-example guided inductive synthesis (CEGIS) loop for better generalization over the state space $\mathcal{D}$. \citet{alfarano2024global} pre-trains a transformer on global Lyapunov function datasets with beam search for candidate generation. SOS methods formulate Lyapunov functions as the feasible solutions of some semi-definite programming tasks and solve these tasks via convex optimization tools. Details of each baseline can be found in Appendix \ref{baseline}.

\begin{table*}[th!]
\centering
\caption{Training time and success rate comparison between ours and the neural baselines. The Succ \% is the successful rate of finding a valid Lyapunov function within 5 hours out of 5 random seeds. Runtime is the average training time for a successful trial. 
}
\label{table:baseline comp}
\vskip 0.15in
\begin{tabular}{l|cc|cc|cc|cc}
\toprule
& \multicolumn{2}{c|}{\textbf{2-D Dynamics}} & \multicolumn{2}{c|}{\textbf{3-D Dynamics}} & \multicolumn{2}{c|}{\textbf{6-D Dynamics}} & \multicolumn{2}{c}{\textbf{8-D Dynamics} (App. \ref{app:8d})}\\
\textbf{Frameworks} & \textbf{Succ \%} & \textbf{Runtime} & \textbf{Succ \%} & \textbf{Runtime} & \textbf{Succ \%} & \textbf{Runtime} & \textbf{Succ \%} & \textbf{Runtime} \\
\midrule
\textbf{Ours} & \textbf{100} & 165s & \textbf{100} & 124s & \textbf{80} & 6290s & \textbf{80} & 14358s \\
\textbf{ANLC} & 53.3 & \textbf{1.409s} & 46.7 & \textbf{63.91s} & 0 & - & 0 & - \\
\textbf{FOSSIL 2.0} & 80 & 7.708s & 66.7 & 221s & 0 & - & 0 & - \\
\bottomrule
\end{tabular}
\end{table*}

\subsection{Performance Analysis}

Table \ref{tab:performance summary} summarizes the runtime, success rate, and discovered Lyapunov functions for a selection of tested nonlinear systems, ranging from 2-D to 10-D, demonstrating the robustness and scalability of our framework. As dimensionality increases, runtime grows with the exponentially expanding search space, reward calculations, SHGO optimization, and genetic programming.

Unlike existing methods that produce neural Lyapunov functions, our framework yields interpretable \emph{analytical} candidates. For example, it correctly identifies the energy function as a valid Lyapunov function for the simple pendulum. Likewise, for the 3-bus power system (Appendix \ref{app:lossless_power}), it discovers the commonly used energy-based storage function for incremental passive systems \citet{weitenberg2018exponential}. 

\emph{Analytical} Lyapunov functions can potentially bypass the need for formal verification. In the 3-D Trig system (Appendix \ref{app: 3-D Trig Dynamics}), over the state space $\mathcal{D} = \{ (x_1, x_2, x_3) \in \mathbb{R}^3 \bm{\mid} |x_i| \leq 1.5, \forall\: i \in \{1,2,3\} \}$, the positive definiteness of the identified Lyapunov function is evident from its formulation. Moreover, the Lie derivative $L_f V = -2x_2^2 - x_3 \sin(2x_3)$ is directly identifiable as non-positive in $\mathcal{D}$, since $x \sin(x) > 0$ for all $x \in (-\pi, \pi)$. By the invariance principle \citet{khalil2002nonlinear}, the discovered function certifies the asymptotic stability of the origin in state space $\mathcal{D}$. When direct identification is non-trivial, SMT solvers can efficiently verify Lyapunov conditions given analytical formulations' simplicity.

\subsection{Newly Discovered Lyapunov Function}
Despite decades of effort in the control community to identify Lyapunov functions, certain stable dynamics still lack a valid Lyapunov function to directly certify their stability. One example is the lossy frequency dynamics in power systems \citet{41298,9449840},
where ``lossy'' refers to the consideration of energy losses in transmission lines. For simplicity, we focus on a 2-bus (4-D) lossy system with the equilibrium point set at the origin, with detailed descriptions provided in Appendix \ref{app:lossy_power}. Using the proposed method, we successfully discover two local Lyapunov functions valid in the considered region $\mathcal{D} = \{(\delta_1, \delta_2, \omega_1, \omega_2) \in \mathbb{R}^{4}  \bm{\mid} |\delta_i| \leq 0.75 \text{ and } |\omega_i| \leq 2 \text{ for } i = 1,2\}$: 
\begin{align*}
    V_{1}(\delta_1, \delta_2, \omega_1, \omega_2) &= \sum_{i=1}^{2} \omega_{1}^{2} + \Big(\sin(\delta_2) - \sin(\delta_1) + \omega_2 \Big)^2, \\
    V_{2}(\delta_1, \delta_2, \omega_1, \omega_2) &= \sum_{i=1}^{2} \omega_{1}^{2} + \Big(\sin(\delta_2) - \sin(\delta_1) -\omega_1 \Big)^2.
\end{align*}
Both functions are formally verified by the SMT solver within the defined state space. To the best of our knowledge, these are the first analytical Lyapunov functions used to certify the local stability of a 2-bus lossy power system.

\subsection{Comparisons with Baselines}

\begin{table*}[th!]
\centering
\caption{Training/solving time of ours and sum-of-squares (SOS) on polynomial systems, averaged over successful trials. The stable regions (local or global) of the considered dynamics are indicated in smaller font.
}
\label{table:sos_poly}
\vskip 0.15in
\begin{tabular}{l|c|c|c|c|c}
\toprule
\textbf{Test Systems}
& \makecell[c]{\textbf{2-D Systems}\\ \scriptsize(App.~\ref{app:van_der_pol}, local; App. \ref{app:meta_1}, global.)}
& \makecell[c]{\textbf{3-D Systems}\\\scriptsize(App. \ref{app:meta_2}, global.)}
& \makecell[c]{\textbf{6-D System}\\ \scriptsize(App.~\ref{subsec: 6d-poly}, local.)}
& \makecell[c]{\textbf{8-D System}\\\scriptsize(App.~\ref{app:8d}, local.)}
& \makecell[c]{\textbf{10-D System}\\ \scriptsize(App.~\ref{app:10d}, local.)}  
\\
\midrule
\textbf{Ours} & 97s  & 108s & \textbf{1667s} & \textbf{14358s} & \textbf{64223s} \\
\textbf{Sum-of-squares} & \textbf{0.765s} & \textbf{1.503s} & - & - & - \\
\bottomrule
\end{tabular}
\end{table*}

\textbf{Neural Lyapunov Function Baselines.} Table \ref{table:baseline comp} compares our success rate and training runtime with two neural Lyapunov function baselines across various test dynamics. For low-dimensional systems, both baselines achieve notably shorter overall training runtime. 
However, to achieve efficient verification, these methods rely on relatively small neural networks, which fail to converge on more challenging tasks (e.g., the simple pendulum and 3-D Trig dynamics involving trigonometric terms). Consequently, both baselines exhibit lower overall success rates than our approach.

As dimensionality grows, the complexity of the Lie derivative of a neural Lyapunov function increases significantly following Definition \ref{def:lie}, creating severe verification bottlenecks for both baselines’ counter-example feedback paradigm. Even networks with fewer than 15 neurons per layer may require hours to finish formal verification. In contrast, our method remains robust up to 6-D and 8-D systems, as its simpler analytical formulations allow numerical verification to efficiently identify violation regions. Moreover, for final verification, our candidates can pass the SMT solver in milliseconds, thanks to term cancellations and algebraic restructuring made possible by their analytical form.

\textbf{Analytical Lyapunov Function Baselines.} 
We evaluated the pre-trained model of \citet{alfarano2024global} on our benchmarks. Trained solely on globally stable systems with fewer than six states, it produced valid Lyapunov functions only for the low-dimensional examples in Appendices \ref{app:van_der_pol}, \ref{app:meta_1}, \ref{app:meta_2}, \& \ref{app:pendulum}, which have global stability guarantees, and failed on every benchmark that is only locally stable.

Table \ref{table:sos_poly} compares the training and solving times of our method and the SOS approach on polynomial systems from Appendix \ref{poly}. For low-dimensional systems, SOS finds valid Lyapunov functions more efficiently, with shorter runtimes. However, it fails to certify stability for the higher-dimensional systems in Appendices \ref{subsec: 6d-poly}, \ref{app:8d}, and \ref{app:10d}, which are only locally stable. While SOS methods can scale to 10-dimensional or higher systems for global stability verification, local stability requires additional constraints on the Lie derivative, which significantly impact scalability. For example, in the 6D polynomial system from Appendix \ref{subsec: 6d-poly}, verifying local stability with a degree-2 polynomial candidate introduces hundreds of symbolic terms in the Lie derivative constraint, rendering the problem intractable for SOSTOOLS. More details are provided in Appendix \ref{app:sos}.

To use SOS on non-polynomial systems without relaxing the dynamics, we follow the recasting scheme of \citet{papachristodoulou_prajna_2005}, replacing each non-polynomial term with auxiliary variables linked by extra (in)equalities. We applied this procedure to the simple pendulum and the 3-D trig system (Appendices \ref{app:pendulum} \& \ref{app: 3-D Trig Dynamics}). SOS did recover Lyapunov certificates in both cases, but at a much higher computational cost than our framework for the 3-D trig example. Three main limitations are revealed during implementation: (i) recasting demands substantial domain expertise and hand-crafted constraints; (ii) added variables and relations greatly increase the computational burden; and (iii) the formulation certifies only stability, not asymptotic stability. For the locally stable 3-D trig system, eliminating the trigonometric terms introduces four auxiliary variables and two equality constraints, pushing the solve time beyond one hour, versus 157s for our method. These limitations undermine SOS’s practicality for high-dimensional, non-polynomial dynamics.

\begin{table}[h!]
    \centering
    \caption{Training/solving time of ours and sum-of-squares (SOS) on two non-polynomial systems, averaged over successful trials.}
    \vskip 0.15in
    \begin{tabular}{l|c|c}
        \toprule 
        \textbf{Frameworks} & \textbf{App.} \ref{app:pendulum} & \textbf{App.} \ref{app: 3-D Trig Dynamics} \\
        \midrule
        \textbf{Ours} &  288s & \textbf{157s} \\
        \textbf{Sum-of-squares} & \textbf{19.58s} & 6163s \\
    \bottomrule
    \end{tabular}
    \label{tab:sos_non_poly}
\end{table}

\vspace{-6pt}
\subsection{Ablation Studies}
\textbf{Risk-Seeking Quantile $\boldsymbol{\alpha}$.} We compare the performance on the 3-D Trig dynamics without GP refinement under $\alpha = 0.1, 0.5, 1$. With $\alpha = 0.1$, the framework achieves steady convergence, lower variance, and the highest success rate of $\mathbf{66.67\%}$, outperforming $\alpha = 0.5$ ($33.33\%$) and the vanilla policy gradient $\alpha = 1$ ($0\%$). These findings confirm the importance of the risk-seeking strategy (Appendix \ref{app: ablation_alpha}). 

\textbf{Verification Comparison.} We compare three verification methods—1) root-finding \citep{feng2024}, 2) the \textit{dreal} SMT solver, and 3) random sampling—against our approach on the 6-D polynomial system. SHGO-based counterexample feedback is an adversarial reward, allowing fast refinement and stronger final guarantees but risking training instability. In contrast, random sampling yields smoother rewards and more stable training, yet it identifies violations less efficiently. We thus blend both methods to balance final performance and training stability. Meanwhile, the SMT solver and root-finding produce mostly mild violations, offering limited optimization guidance (Appendix \ref{app: ablation_verification}).

\textbf{Expert Guidance.} We evaluate four settings on the 6-D polynomial system to assess the impact of GP refinement and expert guidance: 1) transformer only, 2) GP only, 3) transformer $+$ GP refinement, and 4) transformer $+$ GP refinement $+$ expert guidance (ours).
While the transformer only can achieve a 100\% success rate, it requires triple the training time compared to 3) and 4). GP only fails to converge due to its limited understanding of system dynamics. In contrast, GP refinement and expert guidance learning efficiently accelerate transformer parameters' update and enable faster Lyapunov function discovery (Appendix \ref{app: ablation_gp}). 

\vspace{-6pt}
\section{Conclusion}
This work introduces an end-to-end framework for discovering analytical Lyapunov functions for nonlinear dynamical systems. A symbolic transformer, trained with a risk-seeking policy gradient and augmented by genetic programming, proposes candidate expressions; the SHGO global optimizer rapidly verifies them and generates counterexamples during training; and an SMT solver certifies the final Lyapunov candidates. The framework scales to 10-dimensional dynamics and has discovered previously unknown local Lyapunov functions for lossy power system dynamics. It can be extended to other certificate-function discoveries, such as control barrier functions for safety certificates. 

Several promising future research directions emerge from this work. One is efficiently incorporating physical constants, such as the gravitational constant and object mass, which could significantly improve generalization—though directly introducing them as variables may add unnecessary complexity. Another is extending the framework to discover control Lyapunov functions, which are essential for designing stabilizing feedback laws. Theoretical analysis of the framework, including proofs of completeness and convergence, would also be valuable. Finally, since constructing large-scale datasets for local Lyapunov functions remains challenging, using our approach to refine pre-trained models~\cite{alfarano2024global} offers an exciting opportunity to further advance Lyapunov function discovery.

\vspace{-6pt}
\section*{Acknowledgment}
The authors thank Dr. Huan Zhang at the University of Illinois Urbana-Champaign for helpful discussions. J. Feng is supported by the UC–National Laboratory In-Residence Graduate Fellowship L24GF7923. Y. Shi acknowledges support from NSF ECCS-2200692, DOE grant DE-SC0025495, and the Schmidt Sciences AI2050 Early Career Fellowship.

\section*{Impact Statement}
This paper advances learning-based Lyapunov function discovery in control theory by proposing a systematic approach to finding Lyapunov functions for any dynamical system. Once a valid Lyapunov function is found, it guarantees system stability in the state space and offers insights for control engineering. Our method can be applied in energy systems, robotics, transportation, and other control systems. While these improvements have the potential to enhance safety and efficiency, we do not foresee specific negative ethical or societal consequences arising directly from our approach.


\bibliography{example_paper}
\bibliographystyle{temp}

\newpage
\appendix
\onecolumn



\section*{Contents of Supplementary Materials}

\subsection*{Framework Details:}

\begin{itemize}
    \item Appendix \ref{app: transformer}: Symbolic Transformer Model 
    \item Appendix \ref{app:verification}: Global-optimization-based Numerical Verification
    \item Appendix \ref{app:pg}: Risk-seeking Policy Gradient
    \item Appendix \ref{app:gp}: Genetic Programming
\end{itemize}

\subsection*{Baseline Descriptions:}
\begin{itemize}
    \item Appendix \ref{app:anlc}: Augmented Neural Lyapunov Control (ANLC)
    \item Appendix \ref{app:fossil}: FOSSIL 2.0
    \item Appendix \ref{app:meta}: Global Lyapunov Function Discovery by Pre-trained Transformer
    \item Appendix \ref{app:sos}: Sum-of-Squares Methods
\end{itemize}

\subsection*{Tested Dynamical Systems:}

\begin{itemize}
    \item Appendix \ref{poly}: Polynomial Nonlinear Dynamical System

    \item Appendix \ref{nonpoly}: Non-polynomial nonlinear Dynamical System
\end{itemize}

\subsection*{Ablation Studies:}

\begin{itemize}
    \item Appendix \ref{app: ablation_alpha} Ablation Study I - Risk-seeking Policy Gradient

    \item Appendix \ref{app: ablation_verification} Ablation Study II - Global-optimization-based Numerical Verification Process

    \item Appendix \ref{app: ablation_gp} Ablation Study III - Supervised Learning with Expert Guidance Loss

\end{itemize}

\newpage

\textbf{Computation Resources:} All experiments in this work were performed on a workstation with an AMD Ryzen Threadripper 2920X 12-Core Processor, an Nvidia RTX 2080Ti GPU 11 GB, and an Nvidia RTX 2080 GPU 8 GB.

\section{Symbolic Transformer Model}
\label{app: transformer}

We outline the details of our symbolic transformer model, a conditional generator for analytical candidate expression generation, comprising two components: 1) an encoder, and 2) a decoder. $\phi = \{\zeta, \theta\}$ denotes the transformer parameters, where $\zeta$ and $\theta$ denote the parameters of encoder and decoder respectively.

\subsection{Encoder Structure}

\begin{figure*}[t!]
    \centering
    \includegraphics[width=0.95\textwidth]{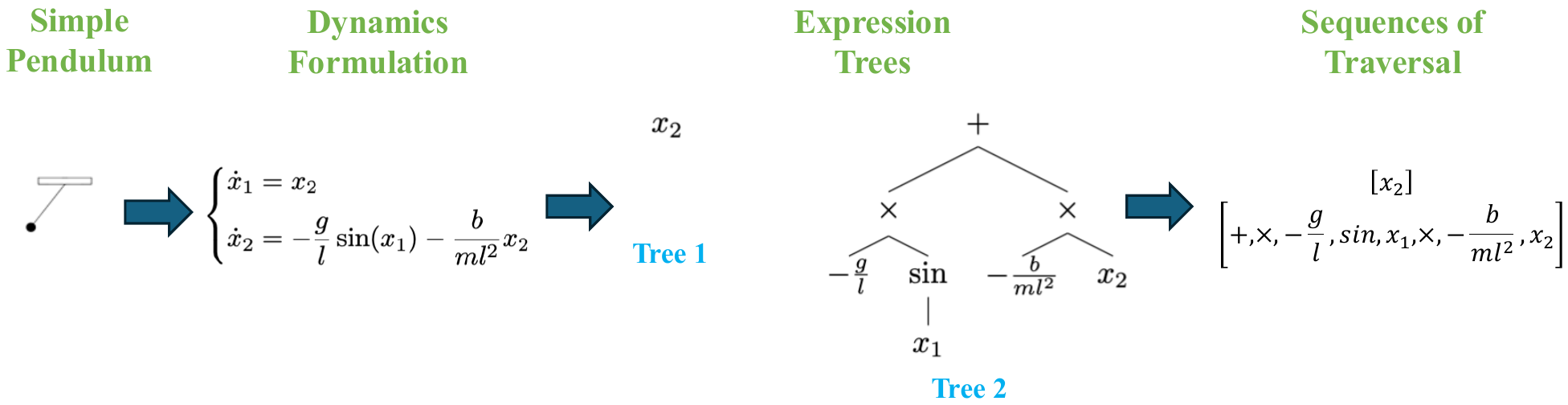}
    \caption{We visualize the dynamics tokenization process of the simple pendulum system. Each analytical formula in the ODE representation of input dynamics is first converted into an expression tree and then represented by the pre-order traversal in symbolic tokens and constant coefficients.}
    \label{fig:pendulum_example}
\end{figure*}

Our framework employs the vanilla transformer encoder from \citet{transformer} to encode two types of input information: 1) the input system dynamics $f(x)$, encoded into a latent vector $\mathcal{F} \in \mathbb{R}^{p}$, where $p \in \mathbb{R}$, and 2) the hierarchical tree state representation \citep{dso} of the selected tokens, encoded as a latent vector $\mathcal{W} \in \mathbb{R}^{k}$, where $k \in \mathbb{R}$. The resulting representations are concatenated as $\mathcal{Z}$. Both inputs are expressed as sequences of symbolic tokens and numerical coefficients when fed into the encoder. For damped pendulum example in Figure \ref{fig:pendulum_example}, suppose we have $m = \SI{0.5}{\kilo\gram}$, $l = \SI{1}{\metre}$, $g = 9.81$, and $b = 0.1$, the dynamics can be tokenized as: $[\texttt{SOS}, x_2, \texttt{EOS}, \texttt{SOS}, +, \times, -, 9, 8, 1, 0, 10^{0}, \sin, x_1, \times, -, 2, 0, 0, 0, 10^{-1}, x_2, \texttt{EOS}]$. In this work, we set the embedding dimension to 128, attention head to 2, and applied a 2-layer transformer encoder for system dynamics $f(x)$ encoding and a 3-layer transformer encoder for hierarchical tree state representation encoding. 

\subsection{Decoder Structure}

The decoder of the symbolic transformer model also uses the vanilla transformer decoder, with an additional linear layer to output the token probability $\psi$ over the symbolic library $\mathcal{L}_{s}$ for token selection. Candidate Lyapunov functions $\Tilde{V}_{\phi}$ are sampled as sequences of symbolic tokens in pre-order traversal. Each symbolic token $\Tilde{V}_{\phi_i}$ is sampled autogressively from conditional distribution $p(\Tilde{V}_{\phi_{i}}|\Tilde{V}_{\phi_{1:(i-1)}},\phi,f(x))$. Upon token sampling is complete for $\Tilde{V}_{\phi}$, we subtract $\Tilde{V}_{\phi}(0)$ from the candidate expression to enforce the Lyapunov condition $\Tilde{V}_{\phi}(0) = 0$. In each epoch, a batch of candidate Lyapunov functions $\{\Tilde{V}_{\phi}^{i} \sim p(\Tilde{V}_{\phi} |\phi, f(x))\}_{i=1}^Q$ is sampled as candidates, which are verified by global-optimization-based numerical verification. In this work, we set the embedding dimension to 128, attention head to 2, and applied a 6-layer transformer decoder for the candidate expression generation. In each epoch, we sample $Q=500$ expressions as candidates.

\section{Global-optimization-based Numerical Verification}
\label{app:verification}

\begin{figure*}[t!]
    \centering
    \includegraphics[width=\textwidth]{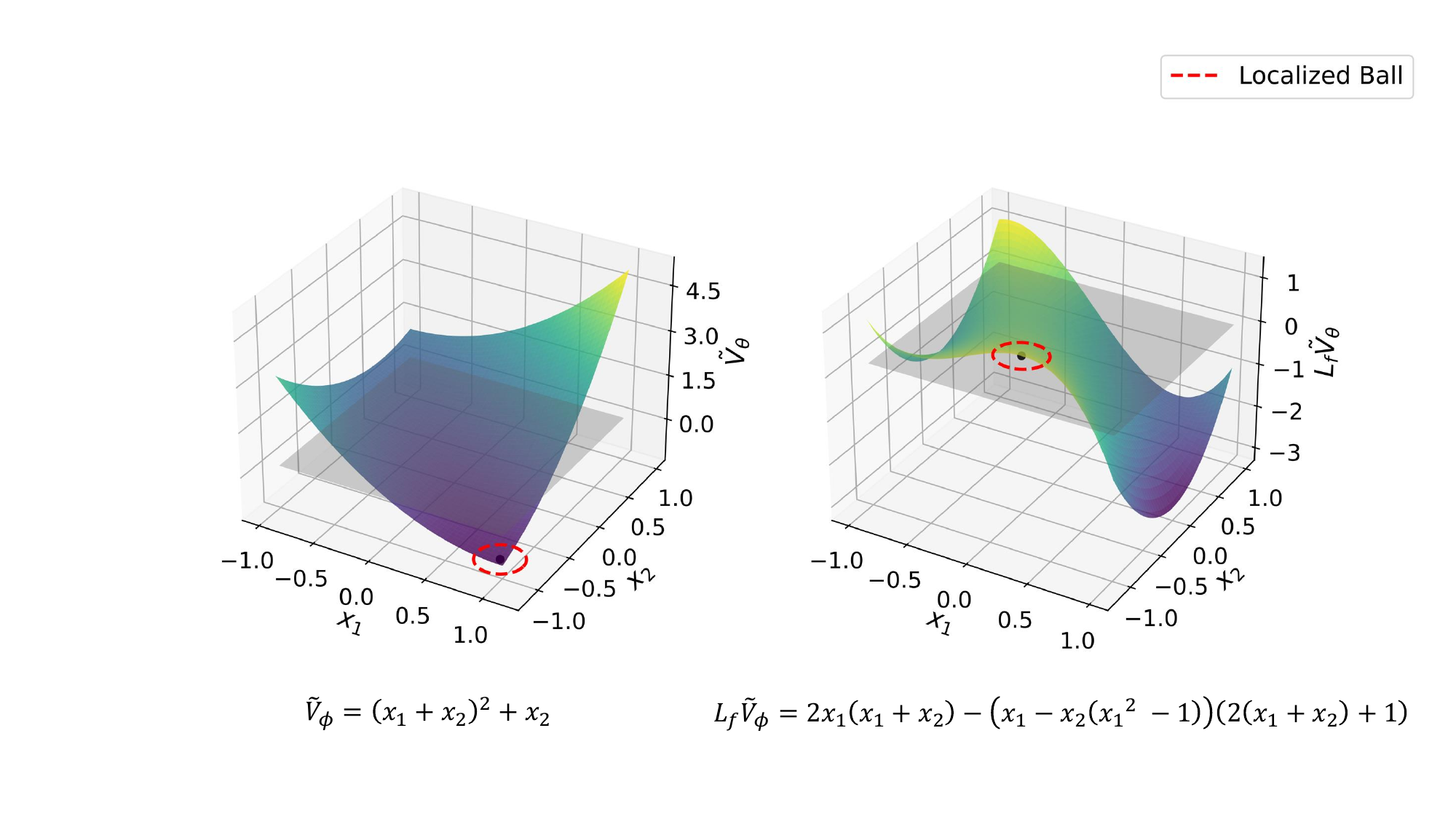}
    \caption{This plot visualizes our proposed verification process on a sampled candidate. $\Tilde{V}_{\phi} = (x_1 + x_2)^2 + x_2$ is a sampled candidate during the training for Van Der Pol Oscillator. Using Simplicial Homology Global Optimization, we first identify the minimizer of the Lyapunov function and the maximizer of the Lie Derivative, the black dots in each graph. Next, data points are sampled in the neighborhoods of the two points, the regions in red circles. For sampled data points that violate the Lyapunov conditions, we feed them into the training set $\mathcal{X}$.}
    \label{fig:shgo_visualization}
\end{figure*}

For a given dynamics $f(x)$, suppose $\Tilde{V}_{\phi}$ is an invalid analytical candidate Lyapunov function. According to Lyapunov conditions defined in Proposition \ref{pro:lyap}, for $x_{1}^{*}, x_{2}^{*} \in \mathcal{D}$, where $x_{1}^{*}, x_{2}^{*}$ are the global minimizers of $\Tilde{V}_{\phi}$ and $-L_{f}\Tilde{V}_{\phi}$ in the state space $\mathcal{D}$, the following two inequalities hold: $\Tilde{V}_{\phi}(x_{1}^{*}) \leq 0$ and $L_{f}\Tilde{V}_{\phi}(x_{2}^{*}) \geq 0$. This implies that if $\Tilde{V}_{\phi}$ is invalid, the neighborhoods of $x_{1}^{*}$ and $x_{2}^{*}$ are highly likely to capture significant violations. Based on this observation, we propose a global-optimization-based numerical verification. This verification identifies minimizers  $x_{1}^{*}$ and $x_{2}^{*}$ by Simplicial Homology Global Optimization (SHGO), verifies Lyapunov conditions on localized samples in neighborhoods $\mathcal{B}_{r}(x_{1}^{*})$ and $\mathcal{B}_{r}(x_{2}^{*})$, and feeds counterexamples back into the training set $\mathcal{X}$. We detail the sampling and condition-checking procedures in Algorithm \ref{alg:verification}. Figure \ref{fig:shgo_visualization} illustrates this verification process on a sampled candidate, $\Tilde{V}_{\phi} = (x_1 + x_2)^2 + x_2$, for the Van der Pol Oscillator. In the implementation, we initiate with $2048$ starting points and iterate $3$ times in the SHGO algorithm for the minimizer detection. We tested the number of starting points with values $[1024, 2048, 4096, 8196]$ on various high-dimensional continuous functions, and the setting with $2048$ starting points achieves the best efficiency. In datapoint sampling for counter-example identification, for each candidate expression, $800$ data points are sampled from each of $\mathcal{B}_{r}(x_{1}^{*})$ and $\mathcal{B}_{r}(x_{2}^{*})$, and additional $800$ data points are randomly sampled across the state space $\mathcal{D}$.  

\begin{algorithm}[h!]
    \caption{Global-optimization-based Numerical Verification}\label{alg:verification}
    \textbf{Input:}  A set of analytical expressions $\mathcal{V} = \{V^{i} |\;i = 1, \cdots, Q\}$, radius $r$, and state space $\mathcal{D}$.\\
    \textbf{Output:} a set of numerically valid candidate $\mathcal{V}^{*}$, a set of encountered counterexample $\mathcal{X}_{ce}$.
    \begin{algorithmic}[1]
        \STATE $\mathcal{V}, \mathcal{X}_{ce} \leftarrow \{\}, \{\}$,
        \FOR{$i=1$ {\bfseries to} $Q$} 
            \STATE $x_{1}^{*}, \; x_{2}^{*} \leftarrow \textit{SHGO}({V}^{i}, \mathcal{D}), \;\textit{SHGO}(-L_{f}{V}^{i}, \mathcal{D})$,
            \COMMENT{Identify global minimizers within the state space $\mathcal{D}$}
            \STATE $\mathcal{X}_1, \mathcal{X}_2, \mathcal{X}_3 \leftarrow \{x_i |\; x_i \: \in \: \mathcal{B}_{r}(x_{1}^{*})\}, \{x_j | \; x_j \: \in \: \mathcal{B}_{r}(x_{2}^{*})\}$, $\{x_k | \; x_k \: \in \: \mathcal{D}\}$
            \STATE Check Lyapunov conditions on $\mathcal{X}_1 \cup \mathcal{X}_2 \cup \mathcal{X}_3$,
            \IF{$R({V}^{i}) = 1$ and no counter example found in $\mathcal{X}_1 \cup \mathcal{X}_2 \cup \mathcal{X}_3$}
                \STATE $\mathcal{V}^{*} \leftarrow \mathcal{V}^{*} \cup \{{V}^{i}\}$.
            \ELSE
                \STATE $\mathcal{X}_{ce} \leftarrow \mathcal{X}_{ce}  \cup \text{identified counterexamples in } \mathcal{X}_1 \cup \mathcal{X}_2 \cup \mathcal{X}_3$. 
                \COMMENT{Gather falsification}
            \ENDIF
        \ENDFOR
        
        \STATE \textbf{Return} $\mathcal{V}^{*}, \mathcal{X}_{ce}$.
    \end{algorithmic}
\end{algorithm}

\section{Risk-seeking Policy Gradient}
\label{app:pg}

\textbf{Objective.}
The standard policy gradient $J_{\text{std}}(\phi) = \mathbb{E}_{\Tilde{V}_{\phi} \sim p(\Tilde{V}_{\phi} | \phi, f(x))} [ R(\Tilde{V}_{\phi})]$ aims to optimize the average performance of a policy given the reward function $R(\cdot)$. However, for the task of Lyapunov function construction, the final performance is measured by identifying a single or a few valid analytical Lyapunov functions that satisfy the Lyapunov conditions. Thus, $J_{\text{std}}(\phi)$ is not an appropriate objective, as there is a mismatch between the objective being optimized and the final performance evaluation metric. To address this misalignment, we adopt risk-seeking policy gradient \citep{dso}, optimizing the best-case performance via the objective $J_{\text{risk}}(\phi, \alpha)$, as defined in Equation \eqref{eq: risk pg}. In implementation, we choose $\alpha = 0.1$ in the training for all tested dynamics.

\begin{proposition}[\citet{dso}]
\label{prop:pg_risk}

Let $J_{\text{risk}}(\phi, \alpha)$ denote the conditional expectation of rewards above the $(1 - \alpha)$-quantile $R_\alpha(\phi)$ as in Equation \eqref{eq: risk pg}. Then the gradient of $J_{\text{risk}}(\phi, \alpha)$ is given by:
\begin{equation}
\label{eq:risk pg gradient}
\nabla_\phi J_{\text{risk}}(\phi, \alpha) = \mathbb{E}_{\Tilde{V}_{\phi} \sim p(\Tilde{V}_{\phi} \mid \phi, f(x))} 
\left[ \left( R_\alpha(\phi) - R(\Tilde{V}_{\phi}) \right) 
\cdot \nabla_\phi \log p(\Tilde{V}_{\phi} \mid \phi, f(x)) \,\middle|\, R(\Tilde{V}_{\phi}) \geq R_\alpha(\phi) \right].
\end{equation}

\end{proposition}

The proposition suggests a Monte Carlo estimate of the gradient of $J_{\text{risk}}(\phi, \alpha)$ from a batch of $N$ samples:
\begin{align}
\label{eq:appro risk pg gradient}
\nabla_\phi J_{\text{risk}} (\phi, \alpha) \approx \frac{1}{\alpha N} \sum_{i=1}^N 
\left[ \tilde{R}_\alpha(\phi) - R(\Tilde{V}_{\phi}^{(i)})  \right] 
\cdot \mathbf{1}_{R(\Tilde{V}_{\phi}^{(i)}) \geq \tilde{R}_\alpha(\phi)} 
\nabla_\phi \log p(\Tilde{V}_{\phi}^{(i)} \mid \phi, f(x)),
\end{align}

where $\tilde{R}_\alpha(\phi)$ is the empirical $(1 - \alpha)$-quantile of the batch of rewards, and $1_{x}$ returns $1$ if condition $x$
is true and $0$ otherwise. Compared to standard REINFORCE algorithm \cite{REINFORCE}, Equation \eqref{eq:appro risk pg gradient} has two distinct features: (1) it has a specific baseline, \(\tilde{R}_\alpha(\phi)\), instead of an arbitrary baseline in standard policy gradients chosen by user; (2) the gradient computation only uses the top $\alpha$ fraction of samples.

\begin{lemma}[\citet{bastiani2024complexityaware}]
\label{lemma:unbounded}
When using the empirical Lyapunov risk in Equation \eqref{eq:risk} as the reward function, the risk-seeking policy gradient is not guaranteed to be unbiased.
\end{lemma}

\begin{proof}
For a given dataset $\mathcal{X} \subseteq \mathcal{D}$, the empirical Lyapunov risk is a random variable whose value is determined by the random expression sampled from the symbolic transformer model. Let $Z$ denote the negated empirical Lyapunov risk, with its probability density $p(z|\phi)$ depending on transformer parameters $\phi$.

$Z$ has range $(-\infty, 0]$. Suppose negated empirical Lyapunov risk is used as the reward function for risk-seeking policy gradient to optimize the average performance of top $(1 - \alpha)$-quantile samples. In Equation \eqref{eq:appro risk pg gradient}, the risk-seeking policy approximates the gradient of the expectation over the truncated random variable $Z_\alpha = Z \cdot \mathbf{1}_{Z\ge s_\alpha}$ with respect to $\phi$, where $s_\alpha$ is the $(1 - \alpha)$-quantile of distribution of $Z$.

\begin{align*}
    s_\alpha = \text{inf}\{ z: \text{CDF}(z) \ge 1 - \alpha \}.
\end{align*}

The probability density of $Z_\alpha$ is given by
\begin{align*}
    p(z_\alpha) = \frac{1}{\alpha} p(z|\phi) \mathbf{1}_{z\ge s_\alpha}.
\end{align*}

To effectively penalize the invalid sampled expressions that miss some state variables or are symbolically invalid, their empirical Lyapunov risks are set to be $\infty$, or a symbolically valid candidate that incorporates all state variables might be less preferable than a symbolically invalid or variable-incomplete expression during the training. Consequently, $s_{\alpha}$ has a non-zero probability density of being $-\infty$. If $s_{\alpha} = -\infty$, then the gradient of the expectation in Equation \eqref{eq:risk pg gradient} is:
\begin{align*}
    \mathbb{E}[Z_\alpha] &= \frac{1}{\alpha} \int_{s_\alpha}^0 z p(z|\phi) \mathrm{d} z, \\
    \mathbb{E}[Z_\alpha] &= \frac{1}{\alpha} \int_{-\infty}^0 z p(z|\phi) \mathrm{d} z, \\
    \nabla_{\phi} \mathbb{E}[Z_\alpha] &= \frac{1}{\alpha} \cdot \nabla_{\phi} \int_{-\infty}^0 z p(z|\phi) \mathrm{d} z.
\end{align*}

Since the integration lower-bound is $-\infty$, the Leibniz rule does not apply, which means interchanging the gradient and the integration changes the result. Therefore, the policy gradient in Equation \eqref{eq:risk pg gradient} is no longer guaranteed to be unbiased.
\end{proof}

\textbf{Reward Design.}
To optimize the symbolic transformer parameters $\phi$ such that the decoder generates a valid candidate Lyapunov function $\Tilde{V}_{\phi}$ satisfying Lyapunov conditions, we employ empirical Lyapunov risk as the fitness metric to measure the violation degree of Lyapunov conditions within the state space following \citet{chang2019neural}. However, as shown in Lemma \ref{lemma:unbounded}, directly using the unbounded empirical Lyapunov risk as the reward for risk-seeking policy gradient might introduce bias. To address this issue, we adopt a bounded reward function using the continuous mapping $g(x) = \frac{1}{x}$ \citep{dso, bastiani2024complexityaware}, defined as: 
\begin{align*}
    R(\Tilde{V}_{\phi}) &= g(\mathcal{L}(\Tilde{V}_{\phi})) 
    =\frac{1}{1 + \mathcal{L}(\Tilde{V}_{\phi})},
\end{align*}
where $\mathcal{L}(\Tilde{V}_{\phi})$ measures the violation degree over the training set \(\mathcal{X}\). This design ensures the reward is bounded in $[0, 1]$, avoiding bias in the risk-seeking policy gradient.

\textbf{Reward Calculation.}
The reward function in Equation \eqref{eq: reward function} quantifies the degree of violation based on empirical data points. However, as the dimensionality of the input dynamics increases linearly, the training set $\mathcal{X}$ must grow exponentially to maintain precision in the empirical measurement of violation degree, which is impractical for high-dimensional cases. To ensure the reward signal remains a reliable indicator of expression quality in the risk-seeking policy gradient without requiring an excessively large dataset, we incorporate Projected Gradient Descent (PGD) \citep{pgd} into the reward calculation process. Prior to evaluating the reward, PGD is employed on all sampled expressions to efficiently identify a set of ``minimizers'' in the state space. These minimizers are computed from a randomly sampled set of starting points through PGD for each sampled expression, and the process can be parallelized on a GPU, enabling efficient computation. Though PGD may converge to local minima, it remains effective in identifying a few counterexamples that violate Lyapunov conditions across invalid candidates. These counter-examples are then added to the training set $\mathcal{X}$ for reward calculation of all sampled expressions in the current epoch and are removed immediately after the calculation. This process leverages PGD to capture violation data points in advance, ensuring the quality of the reward signal without the need for an excessively large dataset.

\section{Genetic Programming}
\label{app:gp}

In the field of symbolic regression, given the large, combinatorial search space, traditional approaches commonly utilize evolutionary algorithms, especially genetic programming (GP) \citep{koza1992genetic}, to retrieve analytical expressions that approximate the output values $y$ given input data $x$. The GP-based symbolic regression operates by evolving the input population of mathematical expressions through evolutionary operations such as selection, crossover, and mutation. A pre-defined fitness metric serves as the objective function to guide the optimization of the population over successive generations. However, for analytical Lyapunov function construction, GP algorithms lack the capability to directly generate Lyapunov functions from the given dynamics and require an initial population that represents potential Lyapunov functions.

As the search space grows exponentially with the expression complexity and the number of states in input dynamics, it is a challenging task even for the symbolic transformer model to search a valid Lyapunov function for complex, high-dimensional systems. Inspired by \citet{rlgp2021}, we incorporate a GP component into the training framework to complement the symbolic transformer model - the symbolic transformer model outputs a well-behaved initial populations of expressions $\Tilde{V}_{\phi}$, which serve as the starting points for the GP component, and GP component refines $\Tilde{V}_{\phi}$ through evolutionary operations to explore the characteristics of Lyapunov functions that might be overlooked by symbolic transformer. The fitness metric for the GP component is the same as the reward function used in the risk-seeking policy gradient. After each refinement, we select an `elite set' of the top-performing refined expressions, $\Tilde{\mathcal{V}}_{gp}$, based on fitness values. These expressions are treated as ground-truth solutions for the transformer decoder, and transformer parameters $\phi$ are optimized through the expert guidance loss introduced in Subsection \ref{subsec: gp}. In the implementation, the size of `elite set' $\Tilde{\mathcal{V}}_{gp}$ is chosen to be $0.1 Q$, where $Q$ is the number of sampled candidate expressions $\Tilde{V}_{\phi}$ in each epoch.

In our framework, we employ three evolutionary operations: mutation, crossover, and selection, within our Genetic Programming component (DEAP \citep{fortin2012deap}). A mutation operator introduces random variations to an expression, such as replacing a subtree of one expression with another randomly generated subtree. A crossover operator exchanges content between two expressions, e.g., by swapping a subtree of one expression with a subtree of another expression, enabling the combination of their features. A selection operator determines which expressions persist into the next population. A common method is tournament selection \citep{koza1992genetic}, where a set of $l$ candidate expressions is randomly sampled from the population, and the expression with the highest fitness value is selected. In each iteration of GP evolution, each expression has a probability of undergoing mutation and a probability of undergoing crossover; selection is performed until the new generation’s population has the same size as the current generation’s population. In empirical experiments, we set the probability of undergoing mutation and crossover to be $0.5$, and we adjust the size of the tournament and number of evolutions proportional to the dimension of the input system.

\begin{algorithm}[t!]
    \caption{Expert Guidance Loss}\label{alg:weighted ce}
    \textbf{Input:}  `Elite set' of analytical expressions $\Tilde{\mathcal{V}}_{gp}$, input dynamics $f(x)$, and transformer parameters $\phi$.\\
    \textbf{Output:} The weighted cross-entropy loss between the transformer model output probability distribution and given refined expressions $\Tilde{\mathcal{V}}_{gp}$.
    \begin{algorithmic}[1]
        \STATE $G \gets |\Tilde{\mathcal{V}}_{gp}|$, 
        \COMMENT{Get the size of `elite set'}
        \STATE $\mathcal{L} \gets 0$,
        \FOR{i = 1 {\bfseries to} $G$}
            \STATE $\mathcal{L} \gets \mathcal{L}+\frac{1}{k_i} R(\Tilde{V}_{gp}^{i}) \sum_{j = 1}^{k_i} -\log \left(p(\Tilde{V}_{gp_j}^{i}|\Tilde{V}_{gp_{1:(j-1)}}^{i},\phi,f(x))\right) $, \COMMENT{Calculate the expert guidance loss based on `elite set' $\Tilde{\mathcal{V}}_{gp}$. Equation \eqref{eq: weighted ce}}
        \ENDFOR
        \STATE $\mathcal{L}(\Tilde{\mathcal{V}}_{gp}) \gets \frac{1}{G} \mathcal{L}$, 
        \STATE \textbf{Return} $\mathcal{L}(\Tilde{\mathcal{V}}_{gp})$.
    \end{algorithmic}
\end{algorithm}

\section{Baseline Descriptions}
\label{baseline}

\subsection{Augmented Neural Lyapunov Control (ANLC)}
\label{app:anlc}

The Augmented Neural Lyapunov Control (ANLC) \citep{ANLC} combines Artificial Neural
Networks (ANNs) with Satisfiability Modulo Theories (SMT) solvers to synthesize stabilizing control laws for the input dynamics $f(x)$ with formal guarantees. The neural network is trained over a dataset of state-space samples to generate candidate control laws and Lyapunov functions, while the SMT solvers are tasked with certifying the Lyapunov conditions of the neural Lyapunov function over a continuous domain and returning a counterexample if the function is invalid. To ease the computationally inefficient verification process in the SMT module, ANLC proposed a discrete falsifier, which discretized the state space for sample selection and evaluation, employed before the SMT call to avoid the frequent calling of the time-consuming SMT falsifier. As the previous learning-based Lyapunov function construction approaches usually initialized the parameters of control policy with pre-computed gains from state-feedback controllers, e.g. Linear-Quadratic Regulators, which requires user time and control expertise to properly perform the initialization process,  ANLC instead removes the need of control initialization by its proposed compositional control architecture containing both linear and nonlinear control laws so that the proposed method allows the synthesis of nonlinear (as well as linear) control laws with the sole requirement being the knowledge of the system dynamics. For empirical experiments, we tested the ANLC algorithm for all system dynamics in Appendices \ref{poly} and \ref{nonpoly}. We tested on the Van Der Pol Oscillator and 3D Trig dynamics to get the best hyperparameter setting. In bold, we show the chosen parameters, selected to have the best success discovery rate on Van Der Pol Oscillator and 3D Trig Dynamics.
\begin{itemize}
    \item lr = $[0.1, \mathbf{0.01}, 0.001]$
    \item activations = $[(x^2, x^2, x^2), (\tanh, \tanh, x^2), \boldsymbol{(x^2, x^2, \tanh)}, (\tanh, \tanh, \tanh)]$
    \item hidden neurons = $[6, \mathbf{12}, 15, 20]$
    \item data = $[500, \mathbf{1000}, 2000]$
    \item iteration = $[500, \mathbf{1000}, 2000]$
\end{itemize}

\subsection{FOSSIL 2.0}
\label{app:fossil}

FOSSIL 2.0 \citep{edwards2024fossil} is a software tool for robust formal synthesis of certificates (e.g., Lyapunov and barrier functions) for dynamical systems modelled as ordinary differential and difference equations. FOSSIL 2.0 implements a counterexample-guided inductive synthesis (CEGIS) for the construction of certificates alongside a feedback control law. In the loop of CEGIS, the learner, based upon neural network templates, 
acts as a candidate to satisfy the conditions over a finite set $\mathcal{D}$ of samples, while the verifier (formal verification tools) works in a symbolic environment that either confirms or falsifies whether the candidate from learner satisfies the conditions over the whole dense domain $\mathcal{X}$. If the verifier falsifies the candidate, one or more counterexamples identified by the verifier are added to the sample set, and the network is retrained. This loop repeats until the verification proves that no counterexamples exist or until a timeout is reached. Similar to the ANLC, in the empirical experiment, we set the hyperparameters based on the Van Der Pol Oscillator and 3-D Trig dynamics and tested for all other dynamics in Appendices \ref{poly} and \ref{nonpoly}.
\begin{itemize}
    \item lr = $[\mathbf{0.1}, 0.01, 0.001]$
    \item activations = $[(x^2, x^2), (\tanh, \tanh, x^2), \boldsymbol{(\tanh, x^2)}]$
    \item hidden neurons = $[\mathbf{6},10, 12]$
    \item data = $[500, \mathbf{1000}, 2000]$
    \item iteration = $[25, \mathbf{50}, 100]$
\end{itemize}

\subsection{Global Lyapunov Function Discovery by Pre-trained Transformer}
\label{app:meta}

\citet{alfarano2024global} pre-trained a transformer on backward-generated and forward-generated global Lyapunov function datasets. The backward-generated datasets involve sampling arbitrary positive definite functions and deriving corresponding stable dynamics through some specific symbolic designs, while the forward-generated polynomial datasets contain randomly generated dynamics with corresponding Lyapunov functions identified by SOS methods if the system is inherently globally stable. Candidate Lyapunov expressions are sampled using beam search in a token-by-token manner. However, their method cannot adaptively refine the candidate Lyapunov functions if the beam search fails on specific dynamics, and it requires a dataset that is expensive to generate (e.g., thousands of CPU hours for a 5-D dynamics dataset) to achieve adequate generalization during inference. Furthermore, its emphasis on global stability limits its applicability to real-world, nonpolynomial control systems, which typically only admit local stability. Due to the lack of resources of multiple industrial-level GPUs, we contacted the authors of \citet{alfarano2024global} to conduct the evaluation of their pre-trained model on our test systems, which is shown in Section \ref{sec:exp}.

\subsection{Sum-of-Squares (SOS) Methods}
\label{app:sos}

SOS methods formulate Lyapunov functions discovery of given dynamics as a semi-definite programming task, where the coefficients of a pre-defined SOS candidate expressions are optimized to satisfy the Lyapunov conditions (hard constraints in the optimization problem) using convex optimization tools. SOS methods are generally applied to polynomial systems for stability analysis. With proper recasting techniques, SOS methods can also be applied to non-polynomial systems.

\begin{definition}[Sum of Squares, \citet{1470374}]

For $x \in \mathbb{R}^n$, a multivariate polynomial $p(x)$ is a sum of squares (SOS) if there exist some polynomials $f_i(x), i = 1,\cdots, M$ such that 

\begin{equation*}
    p(x) = \sum_{i=1}^{M} f_i(x)^2.
\end{equation*}

\end{definition}

\textbf{Polynomial systems.} By \citet{1470374}, for a given $n$-dimensional polynomial dynamics $f(x)$ and an integer degree 2$d$, to check the globally asymptotical stability of $f(x)$, SOS method aims to find a polynomial $V(x)$ of degree $2d$, such that
\begin{enumerate}
    \item $V(x)-\sum_{i=1}^{n}\sum_{j=1}^{d}\epsilon_{ij}x_i^{2j}$ is a SOS, where $\sum_{j=1}^{d}\epsilon_{ij}>\gamma,\forall i = 1,...,n$ with $\gamma>0$, and $\epsilon_{ij}\geq0$ $\forall$ $i$ and $j$,

    \item $-\frac{\partial V}{\partial x}f(x)$ is a SOS. 
\end{enumerate}

For local stability analysis, consider a ball of radius $r$ centered at origin $\mathcal{B}_r(0)$, which can be represented by the semialgebraic set $S=\{x: g(x,r)\geq0$, where $g(x,r)=r-\sum_{i=1}^{n}x_i^2\}$. We require that the stability condition holds in $S$. Retaining the same optimization objective and constraints on $V(x)$ as before, a modified constraint on Lie derivative is imposed: $-\frac{\partial V}{\partial x}f(x)-s(x)g(x,r)$ is a SOS for some SOS $s(x)$. If such an $s(x)$ exists, we can establish local stability. 

In Section \ref{sec:exp}, we develop our code based on the \texttt{findlyap} function from SOSTOOLS (MATLAB) and \href{https://github.com/oxfordcontrol/SOSTOOLS/issues/16}{issue-16} of SOSTOOLS' official GitHub repo to examine the SOS method on polynomial systems in Appendix \ref{poly}. Table \ref{table:sos_poly_full} summarizes the experiment results of the SOS approach on our polynomial test dynamics.

\begin{table}[h!]
    \centering
    \caption{Training\textbackslash solving time of sum-of-squares (SOS) on test polynomial systems.}
    \vskip 0.15in
    \begin{tabular}{l|c|c|c|c|c|c|c}
        \toprule 
        \textbf{Systems} & App. F.1 & App. F.2 & App. F.3 - I & App. F.3 - II & App. F.4 & App. F.5 & App. F.6 \\
        \midrule
        \textbf{Degree 2d} & 2 & 2 & 2 & 4 & 2 & 2 & 2 \\
        \textbf{Region} & $\mathcal{B}_{1}(0)$ & Global & Global & Global & $\mathcal{B}_{1}(0)$ & $\mathcal{B}_{1}(0)$ & $\mathcal{B}_{1}(0)$  \\
        \textbf{Runtime} & 0.697s & 0.832s & 0.497s & 2.509s & - & - & - \\
    \bottomrule
    \end{tabular}
    \label{table:sos_poly_full}
\end{table}

\textbf{Non-Polynomial System.} To apply SOS method on non-polynomial system $\dot{z}=f(z),z\in\mathcal{D}_1$, define $x_1=z$ as the original states and $x_2$ as the newly introduced variables to recast non-polynomial terms in $f(z)$. Let $x=(x_1,x_2)$. The system dynamics can then be written in rational polynomial forms: 
\begin{align*}
\begin{cases}
    \dot{x}_1 &= f_1(x), \\
    \dot{x}_2 &= f_2(x),
\end{cases}
\end{align*}
with constraints $x_2=F(x_1),G_1(x)=0,G_2(x)\geq0,$
where $F,G_1,G_2$ are vectors of functions, to restrict the states of recast dynamics $x = (x_1, x_2)$ equal to the manifold of the original state $z$. 

Define $g(x)$ as the collective denominator of $f_1,f_2$, and local region of interest as a semialgebraic set: 
$$\{(x)\in\mathbb{R}^{n+m}|G_\mathcal{D}(x)\geq0\},$$ where $G_\mathcal{D}(x)$ is a vector of polynomials designed to match the original state space. Let $x_{2,0}=F(0)$. Suppose there exist polynomial functions $V(x)$, $\lambda_1(x),\lambda_2(x)$, and SOS polynomials $\sigma_i(x),i=1,2,3,4$, of appropriate dimensions such that
\begin{align}
    &V(0,x_{2, 0})=0, \label{eq:recast_1} \\
    &V(x)-\lambda_1^T(x)G_1(x)-\sigma_1^T(x)G_2(x)-\sigma_3^T(x)G_D(x) -\phi(x)\in\text{SOS}, \label{eq:recast_2}\\
    &-g(x)(\frac{\partial V}{\partial x_1}(x)f_1(x)+\frac{\partial V}{\partial x_2}(x)f_2(x))-\lambda_2^T(x)G_1(x)-\sigma_2^T(x)G_2(x)-\sigma_4^T(x)G_D(x)\in\text{SOS}, \label{eq:recast_3}
\end{align}

where $\phi(x)$ is some scalar polynomials with $\phi(x_1,F(x_1))>0,\forall x_1\in\mathcal{D}_1\backslash \{0\}$. If constraints \eqref{eq:recast_1}, \eqref{eq:recast_2}, and \eqref{eq:recast_3} hold, then $z=0$ is stable (not necessarily asymptotically stable) \citep{papachristodoulou_prajna_2005}. In Section \ref{sec:exp}, we develop our code from SOSTOOLS to examine the SOS methods on two non-polynomial systems: simple pendulum and 3-D trig dynamics (Appendices \ref{app:pendulum} \& \ref{app: 3-D Trig Dynamics}).

\section{Polynomial Nonlinear Dynamical System}
\label{poly}

\subsection{Van Der Pol Oscillator}\label{app:van_der_pol}
Van Der Pol Oscillator is a nonconservative, oscillating system with nonlinear damping \cite{zhou2022neural}. The dynamics of the Van Der Pol Oscillator have two state variables, formulated as follows:
\begin{align*}
    \Dot{x}_1 &= x_2, \\
    \Dot{x}_2 &= -x_1 - \mu(1 - x_{1}^2) \cdot x_2,
\end{align*}
where $x_1$ and $x_2$ represent the object's position in the Cartesian coordinate, parameter $\mu \in \mathbb{R}^{+}$ indicates the strength of the damping. Under the state space $\mathcal{D} = \{ (x_1, x_2) \in \mathbb{R}^2 \bm{\mid} |x_i| \leq 1\}$ and setting $\mu = 1$, our proposed method found valid local Lyapunov function $V(x_1, x_2) = x_{1}^2 + x_{2}^2$. Other forms of Lyapunov functions for Van Der Pol Oscillator, for example, $V(x_1, x_2) = x_1^2 + x_2(x_1 + x_2)$, are also recovered during the experiments.

\subsection{Two-variable-polynomial-system with higher degree}
\label{app:meta_1}

Here we have a polynomial system of two variables with a higher degree, adopted from \citet{alfarano2024global}, formulated as:

\begin{align*}
\begin{cases}
    \Dot{x}_1 &= -5x_1^3 - 2x_1 \cdot x_2^2, \\
    \Dot{x}_2 &= -9x_1^4 + 3x_1^3 \cdot  x_2 - 4x_2^3.
\end{cases}
\end{align*}

Under the state space $\mathcal{D} = \{ (x_1, x_2) \in \mathbb{R}^2 \bm{\mid} |x_i| \leq 1\}$, our proposed method successfully found valid local Lyapunov function $V(x_1, x_2) =  9x_1^2 + x_2^2$.

\subsection{Three-variable-polynomial-systems with higher degree}
\label{app:meta_2}

Table \ref{table:3d_poly_h} describes two polynomial systems of three variables with a higher degree, adopted from \citet{alfarano2024global}. Our framework successfully retrieves valid local Lyapunov functions on both examples under the state space $\mathcal{D} = \{ (x_1, x_2, x_3) \in \mathbb{R}^3 \bm{\mid} |x_i| \leq 1\}$.

\begin{table}[h]
\caption{Three-Dimensional Polynomial Example with Higher Degree.}
\vskip 0.15in
\centering
\begin{tabular}{l|l}
\toprule
\textbf{System} & \textbf{Lyapunov function} \\
\midrule
$\begin{cases} 
\Dot{x}_1 = -3x_1^3+3x_1 \cdot  x_3-9x_1\\ 
\Dot{x}_2 = -x_1^3-5x_2+5x_3^2\\
\Dot{x}_3 = -9x_3^3
\end{cases}$
& $V(x_1, x_2, x_3) = 9x_1^2 + x_2^2 + x_3^2$ \\
\midrule
$\begin{cases} 
\dot{x}_1 = -8x_1 \cdot x_2^2-10x_2^4\\ 
\dot{x}_2 = -8x_2^3+3x_2^2-8x_2\\
\dot{x}_3 = -x_3
\end{cases}$
& $V(x_1, x_2, x_3) = x_1^8 \cdot x_2^2 \cdot x_3^2 + x_2^2 $ \\
\bottomrule
\end{tabular}
\label{table:3d_poly_h}
\end{table}

\subsection{6-D Polynomial Nonlinear System}
\label{subsec: 6d-poly}

This 6-D dynamics consists of three two-dimensional asymptotically stable linear subsystems that are coupled by three nonlinearities with small gains adopted from \citet{6ddynamics}. The dynamics are written as: 
\begin{align*}
    \Dot{x}_1 &= - x_1 + 0.5 x_2 - 0.1 x_{5}^{2}, \\
    \Dot{x}_2 &= -0.5 x_1 - x_2, \\
    \Dot{x}_3 &= - x_3 + 0.5 x_4 - 0.1 x_{1}^{2}, \\
    \Dot{x}_4 &= -0.5 x_3 - x_4, \\
    \Dot{x}_5 &= - x_5 + 0.5 x_6, \\
    \Dot{x}_6 &= -0.5 x_5 - x_6 + 0.1 x_{2}^{2}.
\end{align*}
    
Our proposed method is trained over the state space $\mathcal{D} = \{ (x_1, x_2, x_3, x_4, x_5, x_6) \in \mathbb{R}^6 \bm{\mid} |x_i| \leq 1, \forall i \in \{1,2,..,6\}\}$, and is able to find a valid Lyapunov function $V(x) = x_{1}^{2} + x_{2}^{2} + x_{3}^{2} + x_{4}^{2} + x_{5}^{2} + x_{6}^{2}$. This dynamics is not globally asymptotically stable since if $x_1, x_2, \text{ or } x_5$ has a significantly large value, the perturbations introduced by the small gains will shift the object by a significant amount away from the equilibrium point. By empirical checking, our found Lyapunov function certifies the asymptotical stability of this system over the region $\mathcal{D}' = \{ (x_1, x_2, x_3, x_4, x_5, x_6) \in \mathbb{R}^6 \bm{\mid} \sum_{i=1}^{6} x_i^2 \leq 500\}$.


\subsection{8-D Polynomial Nonlinear System}\label{app:8d}

This 8-D dynamics consists of four two-dimensional asymptotically stable linear subsystems that are coupled by four nonlinearities with small gains, modified from the above 6D polynomial dynamics. The dynamics are written as: 
\begin{align*}
    \Dot{x}_1 &= - x_1 + 0.5 x_2 - 0.1 x_{5}^{2}, \\
    \Dot{x}_2 &= -0.5 x_1 - x_2, \\
    \Dot{x}_3 &= - x_3 + 0.5 x_4 - 0.1 x_{1}^{2}, \\
    \Dot{x}_4 &= -0.5 x_3 - x_4, \\
    \Dot{x}_5 &= - x_5 + 0.5 x_6 + 0.1 x_7^2, \\
    \Dot{x}_6 &= -0.5 x_5 - x_6, \\
    \Dot{x}_7 &= - x_7 + 0.5 x_8, \\
    \Dot{x}_8 &= -0.5 x_7 - x_8 - 0.1 x_4^2. \\
\end{align*}
    
Our proposed method is trained over state space $\mathcal{D} = \{ (x_1, x_2, x_3, x_4, x_5, x_6, x_7, x_8) \in \mathbb{R}^8 \bm{\mid} |x_i| \leq 1, \forall i \in \{1,2,..,8\}\}$, and is able to find a valid Lyapunov function $V(x) = x_{1}^{2} + x_{2}^{2} + x_{3}^{2} + x_{4}^{2} + x_{5}^{2} + x_{6}^{2}+ x_{7}^{2}+ x_{8}^{2}$. This Lyapunov function certifies the asymptotical stability of this system over the region $\mathcal{D} = \{ (x_1, x_2, x_3, x_4, x_5, x_6, x_7, x_8) \in \mathbb{R}^8 \bm{\mid} \sum_{i=1}^{8} x_i^2 \leq 450\}$. This dynamics is not globally asymptotically stable since if $x_1, x_4, x_5, \text{ or } x_7$ has a significantly large value, the perturbations introduced by the small gains will shift the object by a significant amount away from the equilibrium point.

\subsection{10-D Polynomial Nonlinear System}\label{app:10d}

Finally, we extend to the original 10-D polynomial dynamics proposed in \citet{6ddynamics}. This 10-D dynamics consists of five two-dimensional asymptotically stable linear subsystems that are coupled by four nonlinearities with small gains. The dynamics are written as: 
\begin{align*}
    \Dot{x}_1 &= - x_1 + 0.5 x_2 - 0.1 x_{5}^{2}, \\
    \Dot{x}_2 &= -0.5 x_1 - x_2, \\
    \Dot{x}_3 &= - x_3 + 0.5 x_4 - 0.1 x_{1}^{2}, \\
    \Dot{x}_4 &= -0.5 x_3 - x_4, \\
    \Dot{x}_5 &= - x_5 + 0.5 x_6 + 0.1 x_{9}^2, \\
    \Dot{x}_6 &= -0.5 x_5 - x_6, \\
    \Dot{x}_7 &= - x_7 + 0.5 x_8, \\
    \Dot{x}_8 &= -0.5 x_7 - x_8. \\
    \Dot{x}_9 &= - x_9 + 0.5 x_{10}, \\
    \Dot{x}_{10} &= -0.5 x_9 - x_{10} - 0.1 x_{4}^{2}. \\
\end{align*}

Our proposed method is trained over the state space $\mathcal{D} = \{ (x_1, x_2, x_3, x_4, x_5, x_6, x_7, x_8, x_9, x_{10}) \in \mathbb{R}^{10} \bm{\mid} |x_i| \leq 1, \forall i \in \{1,2,..,10\}\}$, and is able to find a valid Lyapunov function $V(x) = x_{1}^{2} + x_{2}^{2} + x_{3}^{2} + x_{4}^{2} + x_{5}^{2} + x_{6}^{2}+ x_{7}^{2}+ x_{8}^{2}+ x_{9}^{2}+ x_{10}^{2}$. This Lyapunov function certifies the asymptotic stability of this system over the region $\mathcal{D} = \{ (x_1, x_2, x_3, x_4, x_5, x_6, x_7, x_8, x_9, x_{10}) \in \mathbb{R}^{10} \bm{\mid} \sum_{i=1}^{10} x_i^2 \leq 400\}$. This dynamics is not globally asymptotically stable since if $x_1, x_4, x_5, \text{ or } x_9$ has a significantly large value, the perturbations introduced by the small gains will shift the object by a significant amount away from the equilibrium point.

\section{Non-polynomial Nonlinear Dynamical Systems}
\label{nonpoly}

\subsection{Simple Pendulum}\label{app:pendulum}

The simple pendulum is a well-known classical nonlinear system that contains two state variables. The dynamics are formulated as follows,
\begin{align*}
    \Dot{x}_1 &= x_2, \\
    \Dot{x}_2 &= - \frac{g}{l} \sin(x_1) - \frac{b}{m}x_2,
\end{align*}
where $x_1$ is the angular position from the inverted position, $x_2$ is the angular velocity, and parameters $g, m, l, b$ are the acceleration of gravity, the mass of the inverted object, the length of the string, and the coefficient of friction, respectively. In experiments, since we don't incorporate the constant generation capability within the training framework, we set $g = 1$, $m = \SI{1}{\kilo\gram}$, $l = \SI{1}{\metre}$, and $b = 0.1$. Our proposed method finds the valid Lyapunov function $V = 2 - 2\cos(x_1) +x_{2}^2$ over the state space: $\mathcal{D} = \{ (x_1, x_2) \in \mathbb{R}^2 \bm{\mid} |x_1| \leq \pi \text{ and } |x_2| \leq 6 \}$. This found Lyapunov function has the same analytical structure as the energy function of the inverted pendulum. 

\subsection{3-D Trigonometric System}
\label{app: 3-D Trig Dynamics}
3-D trig dynamics comes from exercise problems in textbook \citet{khalil2002nonlinear} whose  dynamics are written as follows,
\begin{align*}
    \Dot{x}_1 &= x_2, \\
    \Dot{x}_2 &= -h(x_1) - x_2 - h(x_3), \\
    \Dot{x_3} &= x_2 - x_3,
\end{align*}
where $h(x) = \sin(x) \cdot \cos(x)$. When the state space is $\mathcal{D} = \{ (x_1, x_2, x_3) \in \mathbb{R}^3 \bm{\mid} |x_i| \leq 1.5, \forall \; i \in\{1,2,3\} \}$, the valid local Lyapunov function found by our proposed method is $V(x_1, x_2, x_3) = 1 - \cos(x_1)^2 + x_2^2 + \sin(x_3)^2$, which is consistent to the textbook solution of Lyapunov function for this particular dynamics.

\subsection{N-bus Lossless Power System}
\label{app:lossless_power}
We test our proposed framework on the N-bus power lossless system \citep{cuipower,10820007} to examine its ability to handle complex high-dimensional dynamics. Consider $\theta_i$, $\omega_i$ as the phase angle and the frequency of bus $i$, respectively, the dynamics for each bus are formulated as follows,
\begin{align*}
   \Dot{\theta_{i}} &= \omega_{i}, \\
    m_{i} \Dot{\omega}_{i} &= p_{i} - d_i \omega_i - u_i(\omega_i) - \sum_{j = 1}^{N} B_{ij} \cdot \sin(\theta_i - \theta_j),
\end{align*}
where $m_i$ is the generator inertia constant, $d_i$ is the combined frequency response coefficient from synchronous
generators and frequency sensitive load, and $p_i$ is the net power injection, for each bus $i = 1, \cdots, N$. $B \in \mathbb{R}^{N \times N}$ is the susceptance matrix with $B_{ij} = 0 $ for every pair $\{i, j\}$ such that bus $i$ and bus $j$ are not connected, and $u_i(\omega_i)$ is the controller at bus $i$ that adjusts the power injection to stabilize the frequency.

Since the frequency dynamics of the system depends
only on the phase angle differences, so we change the coordinates:
$$\delta_i = \theta_i - \frac{1}{N}\sum_{i = 1}^{N} \theta_i$$
where $\delta_i$ can be understood as the center-of-inertia coordinates of each bus. In our experiment, we test the proposed framework on the 3-bus power system. For simplicity, we set $p_i = 0$, $m_i = 2$, $d_i = 1$, $u_i(\omega_i) = \omega_i$, and $B_{ij} = 1 \: \forall \: i \neq j, B_{ii} = 0$. In this case, the equilibrium point for our system is at the origin, i.e. $\delta_i^{*} =\omega_i^*= 0, \: i = 1, 2, 3$. The state space for our experiment is defined as: $\mathcal{D} = \{ (\delta_1,\delta_2,\delta_3, \omega_1,\omega_2,\omega_3) \in \mathbb{R}^6 \bm{\mid} |\delta_i| \leq 0.75 \text{ and } |\omega_i| \leq 1.2 \text{ for } i = 1, 2, 3\}$. Through our method, we retrieved a valid Lyapunov function $V(\delta_1, \delta_2, \delta_3, \omega_1, \omega_2, \omega_3) = (\sum_{i=1}^{3} \omega_i^2) - 0.5\left(\sum\limits_{i=1}^{3} \sum\limits_{\substack{j=1, i\neq j}}^{3} \cos(\delta_i - \delta_j) - 1 \right)$, which is consistent to the known Lyapunov function presented in \cite{cuipower}. The Lie derivative of the identified Lyapunov function can be simplified as $L_{f}V = -2(\omega_{1}^{2} + \omega_{2}^{2} + \omega_{3}^{2})$. The analytical structure of this found Lyapunov function and invariance principle allows us to easily identify it as a valid Lyapunov function by hand.

\subsection{Indoor Micro Quadrotor} \label{app:quadrotor}

For the angular rotations subsystems of the quadrator from \citet{1302409}, it has 6 states to describe the angular motion of the quadrotor. The states $x_1, x_3, \text{ and } x_5$ describe the roll, pitch, and yaw of the quadrator, and states $x_2, x_4, \text{ and } x_6$ represent their time derivatives. With perturbation terms $\Omega$ and control inputs $U_1, U_2, \text{ and } U_3$, the subsystem can be formulated as follows,
\begin{align*}
    \Dot{x}_1 &= x_2, \\
    \Dot{x}_2 &= x_4 x_6 (\frac{I_y - I_z}{I_x}) - \frac{J_{R}}{I_x}x_4 \Omega + \frac{l}{I_x} U_1, \\
    \Dot{x}_3 &= x_4, \\
    \Dot{x}_4 &= x_2 x_6 (\frac{I_z - I_x}{I_y}) + \frac{J_{R}}{I_y} x_2 \Omega + \frac{l}{I_y}U_2, \\
    \Dot{x}_5 &= x_6, \\
    \Dot{x}_6 &= x_2 x_4 (\frac{I_x - I_y}{I_z}) + \frac{l}{I_z} U_3,
\end{align*}

where $I_x, I_y, \text{ and } I_z$ represents the body inertia, $l$ denotes the lever, and $J_{R}$ is the rotor inertia. With the control policy
\begin{align*}
    U_1 &= -\frac{I_x}{l}(x_1 - x_{1}^{d}) - k_1 x_2,  \\
    U_2 &= -\frac{I_y}{l}(x_3 - x_{3}^{d}) - k_2 x_4,  \\
    U_3 &= - I_z (x_5 - x_{5}^{d}) - k_3 x_6,  
\end{align*}

and restricting $I_x = I_y$, the angular rotations subsystems is stabilized to the chosen equilibrium point $X_{d} = \{x_{1}^{d}, 0, x_{3}^{d}, 0, x_{5}^{d}, 0\}$. In empirical experiments, we set $X_{d} = \{0, 0, 0, 0, 0, 0\}$, state space $\mathcal{D} = \{ (x_1, x_2, x_3, x_4, x_5, x_6) \in \mathbb{R}^6 \bm{\mid} |x_i| \leq 3, \forall i \in \{1,2,..,6\}\}$, $I_x = I_y = 2$, $I_z = 5$, $k_1 = 5$,
$k_2 = 20$, $k_3 = 4$, $l=1$,  $J_{R} = 1$, and $\Omega= \sin(x_2)\cos(x_4)$, our framework successfully found Lyapunov function $V(x_1, x_2, x_3, x_4, x_5, x_6) = \sum_{i=1}^{6} x_{i}^{2}$. By examining the Lie derivative $L_{f}V = -2.5 x_{2}^2 - 10 x_{4}^{2} - 0.8 x_{6}^{2}$ and using the Invariance principle, we conclude that this subsystem is globally asymptotically stable.
Under other parameter settings, our framework can also retrieve valid analytical Lyapunov functions for this dynamics.

\subsection{N-bus Lossy Power System}
\label{app:lossy_power}

Unlike N-bus lossless power systems in Appendix \ref{app:lossless_power} which has a well-known energy-based storage function served as a valid Lyapunov function for asymptotical stability guarantee, the N-bus lossy power system \citep{9449840} does not have a known analytical Lyapunov function to certify stability, though by passivity the system should be asymptotically stable at origin. Utilizing the proposed framework, we aim to discover a valid analytical Lyapunov function for an N-bus lossy power system to formally certify its asymptotic stability.

The $\theta_{i}$ and $\delta_i$ are the angle and frequency deviation of bus $i$, the dynamics of an N-bus lossy power system is represented by the swing equation, formulated as:
\begin{align*}
   \Dot{\theta_{i}} &= \omega_{i}, \\
    m_{i} \Dot{\omega}_{i} &= p_{i} - d_i \omega_i - u_i(\omega_i) - \sum_{j = 1}^{N} B_{ij} \cdot \sin(\theta_i - \theta_j) - \sum_{j = 1}^{N} G_{ij} \cdot \cos(\theta_i - \theta_j),
\end{align*}

where $m_i$ is the generator inertia constant, $d_i$ is the combined frequency response coefficient from synchronous generators and frequency-sensitive load, and $p_i$ is the net power injection, for each bus $i = 1, \cdots , N$. The susceptance and conductance of the line $(i, j)$ are $B_{ij} = B_{ji}$ and $G_{ij} = G_{ji}$, respectively. The value is 0 if the buses are not connected. In this work, we consider input $u_i$ to be a static feedback controller where only its local frequency measurement $\omega_i$ is available. Like the lossless power system, since the frequency dynamics of the system depends only on the phase angle differences, we change the coordinates:
$$\delta_i = \theta_i - \frac{1}{N}\sum_{i = 1}^{N} \theta_i$$
where $\delta_i$ can be understood as the center-of-inertia coordinates of each bus.

We test the proposed framework on a 2-bus lossy power system. In experiment, we set $p_i = 1$,$m_i = 2$, $d_i = 1$, $u_i(w_i) = w_i$, $B_{ij} = 1, G_{ij} = 1 \; \forall \; i\neq j$, $B_{ii} = 0, G_{ij} = 0$. By this setting, the equilibrium point for this system is at the origin, i.e. $\delta_{i}^{*} = \omega_{i}^{*} = 0$. The state space for the experiment is defined as: $\mathcal{D} = \{(\delta_1, \delta_2, \omega_1, \omega_2) \in \mathbb{R}^{4}  \bm{\mid} |\delta_i| \leq 0.75 \text{ and } |\omega_i| \leq 2 \text{ for } i = 1,2\}$. The proposed method found two valid Lyapunov function $V(\delta_1, \delta_2, \omega_1, \omega_2) = \omega_{1}^{2} + \omega_{2}^{2} + (\omega_2 - \sin(\delta_1) + \sin(\delta_2))^2$ and $V(\delta_1, \delta_2, \omega_1, \omega_2) = \omega_{1}^{2} + \omega_{2}^{2} + (-\omega_1 - \sin(\delta_1) + \sin(\delta_2))^2$. Both Lyapunov functions pass the formal verification by SMT solver in the state space $\mathcal{D}\backslash \mathcal{B}_{\epsilon}(0)$, where precision $\delta$ is set to be $e^{-12}$ and $\epsilon = e^{-3}$ to avoid tolerable numerical error.

\subsection{9-D Synthetic Dynamics}
\label{app:9-d_synthetic}

Consider the synthetic dynamics adapted from Appendices \ref{subsec: 6d-poly} \& \ref{app: 3-D Trig Dynamics} with linear interactions between two subsystems:
\begin{align*}
    \dot{x}_1 &= - x_1+0.5x_2-0.1x_5^2, \\
    \dot{x}_2 &= -0.5x_1-x_2+0.1x_8, \\
    \dot{x}_3 &= -x_3+0.5x_4-0.1x_1^2, \\
    \dot{x}_4 &= -0.5x_3-x_4,\\
    \dot{x}_5 &= -x_5+0.5x_6,\\
    \dot{x}_6 &= -0.5x_5-x_6+0.1x_2^2,\\
    \dot{x}_7 &= x_8,\\
    \dot{x}_8 &=-\sin(x_7)\cos(x_7)-x_8-\sin(x_9)\cos(x_9)-0.1x_2,\\
    \dot{x}_9 &= x_8-x_9.
\end{align*}

To properly address the trigonometric terms in $\dot{x}_8$, the Lyapunov function for this dynamics can't be a simple form like $\sum_{i=1}^{n}x_i^2$ and should include some trigonometric terms. Setting the state space $\mathcal{D}=\{x \in \mathbb{R}^9 | |x_i| \leq 1.5, \forall \; i=1,\cdots, 9\}$, our method successfully identifies a valid Lyapunov function $V=\sum_{i=1}^6x_i^2+\sin(x_7)^2+x_8^2-\cos(x_9)+1$, which passes formal verification following settings in Section \ref{sec:exp}.

\section{Ablation Studies}
\label{app: ablation studies}

\subsection{Ablation Study I - Risk-seeking Policy Gradient}
\label{app: ablation_alpha}

\begin{figure*}
    \centering
    \begin{minipage}[t]{0.48\textwidth}
        \centering
        \includegraphics[width=\linewidth]{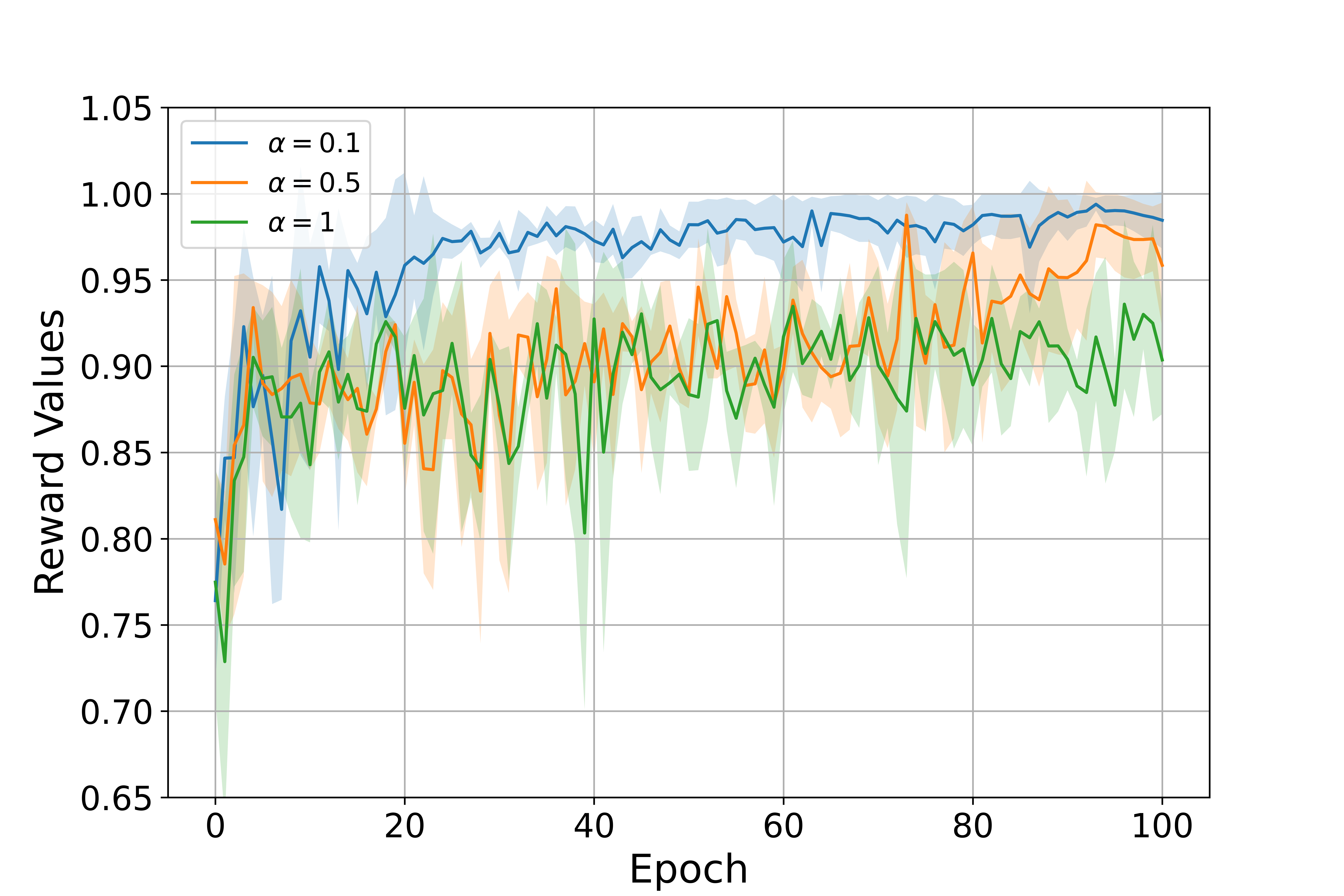}
    \end{minipage}
    \hfill
    \begin{minipage}[t]{0.48\textwidth}
        \centering
        \includegraphics[width=\linewidth]{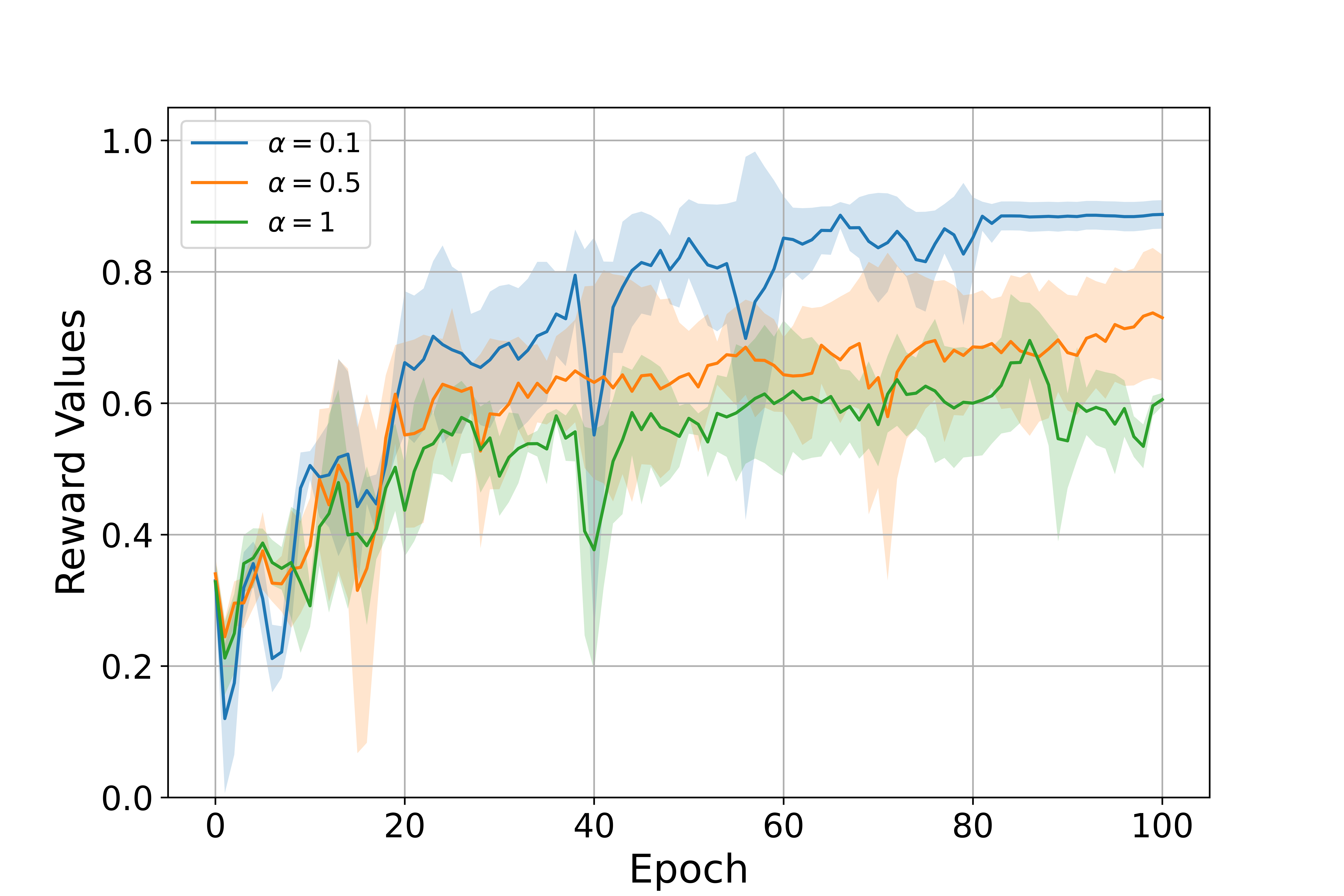}
    \end{minipage}
    \caption {\textbf{Reward Trajectory Comparison for Ablation Study I}: Test proposed framework (without GP component) on 3-D Trig dynamics with three different choices of $\alpha$ for 100 epochs. The figure on the left visualizes the highest reward in the batch of samples obtained in each epoch, and the figure on the right visualizes the $90\%$ quantile of reward distributions of the batch of samples obtained in each epoch.}
    \label{fig: full_reward_traj}
\end{figure*}

To better understand the influence of $\alpha$ in policy gradient on the training performance of the symbolic transformer, we experiment with our framework without GP refinement on 3-D Trig dynamics under different choices of $\alpha$ (the risk-seeking quantile) over 100 epochs. Specifically, we choose $\alpha = 0.1, 0.5, 1$ and compare the transformer model's performance by the reward trajectories and success discovery rate. Figure \ref{fig: full_reward_traj} plots the reward trajectories—both the highest reward and the $90\%$-quantile—under the three different choices of $\alpha$ (the risk-seeking quantile). For $\alpha = 0.1$, the setting we used on tested dynamics in experiments, the highest reward and the 90\% quantile converge steadily to 1 and 0.9, respectively, exhibiting lower variance and faster stabilization. For $\alpha = 0.5$, the median-based threshold increases steadily but shows greater instability. Finally, for $\alpha = 1$ vanilla policy gradient, the best reward reaches only 0.94, while the 90\% quantile hovers around 0.6. Additionally, with $\alpha = 0.1$, the framework achieves the highest success recovery rate ($66.67\%$) compared to $33.33\%$ and $0\%$ for $\alpha = 0.5$ and $\alpha = 1$. These findings correspond to the analysis of misaligned objective between the standard policy gradient and Lyapunov function construction in Subsection \ref{subsec:risk-seeking}, confirm the importance of risk-seeking policy optimization, and highlight the effectiveness of setting $\alpha=0.1$ in experiments.

\subsection{Ablation Study II - Global-optimization-based Numerical Verification Process}
\label{app: ablation_verification}

This work proposes a global-optimization-based numerical verification for candidate verification and counterexamples' feedback within the training paradigm. To identify the effectiveness of the proposed verification algorithm, this ablation study compares the following verification methods on the proposed framework without the GP component on the 6-D polynomial dynamics (Appendix \ref{subsec: 6d-poly}):

\begin{itemize}
    \item \textbf{Global optimization (SHGO):} Proposed framework with global-optimization-based numerical verification in Subsection \ref{sub:ver}.

    \item \textbf{Root finding:} Proposed framework with root-finding numerical verification introduced in \citet{feng2024}.

    \item \textbf{SMT:} Proposed framework with formal verification \textit{dreal} \citep{dReal} SMT solver. 

    \item \textbf{Ramdom sampling:} Proposed framework with random sampling verification.
    
\end{itemize}

\begin{table*}[t!]
\centering
\caption{Number of epoch, training time, success rate, and formal verification time comparison between four different settings in Ablation study II. Runtime is the average training time for successful trials. The Succ. \% is the success rate of finding a valid Lyapunov function out of 3 random seeds. Ver. Time is the average formal verification time of identified candidates on successful trials. The experiment terminates either when a valid Lyapunov function is found (passes formal verification), the training epoch exceeds 250, or a timeout (4 hours limit).}
\vskip 0.15in
\begin{tabular}{c|c|c|c|c}
\toprule
\multicolumn{1}{c|}{\textbf{Stats.}} & \multicolumn{1}{c|}{\textbf{Global optimization}} & \multicolumn{1}{c|}{\textbf{Root finding}} & 
\multicolumn{1}{c|}{\textbf{SMT}} & \multicolumn{1}{c}{\textbf{Random sampling}} \\
\midrule
\textbf{Epoch} & 168 & - & - & \textbf{160} \\
\textbf{Runtime} & 8399s & - & - & \textbf{5684s}  \\
\textbf{Succ. \%} & \textbf{100} & 0 & 0 & \textbf{100} \\
\textbf{Ver. Time} & \textbf{1.67s} & - & - & 15.93s\\
\bottomrule
\end{tabular}
\label{table:ablation-verification}
\end{table*}

\begin{figure*}
    \centering
    \begin{minipage}[th!]{0.48\textwidth}
        \centering
        \includegraphics[width=\linewidth]{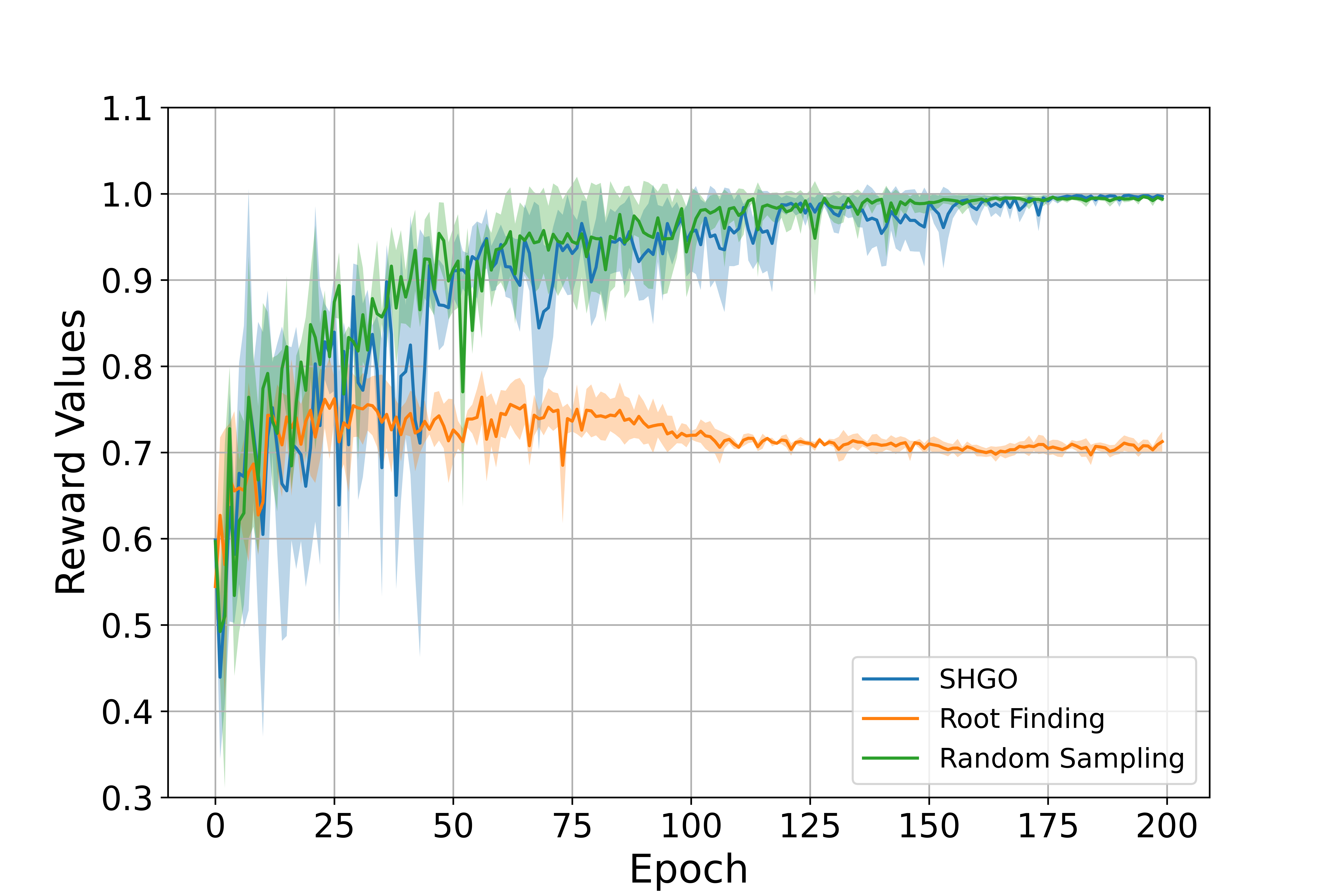}
        \caption{\textbf{Best Reward Trajectory for Ablation Study II}: Visualize the best reward trajectory of our framework with SHGO, root finding, and random sampling schemes in Ablation Study II on 6-D polynomial dynamics. The SMT-verification-based trajectory is not included as it timeouts at the beginning training stage. }
         \label{fig:ablation-2}
    \end{minipage}
    \hfill
    \begin{minipage}[th!]{0.48\textwidth}
        \centering
        \includegraphics[width=\linewidth]{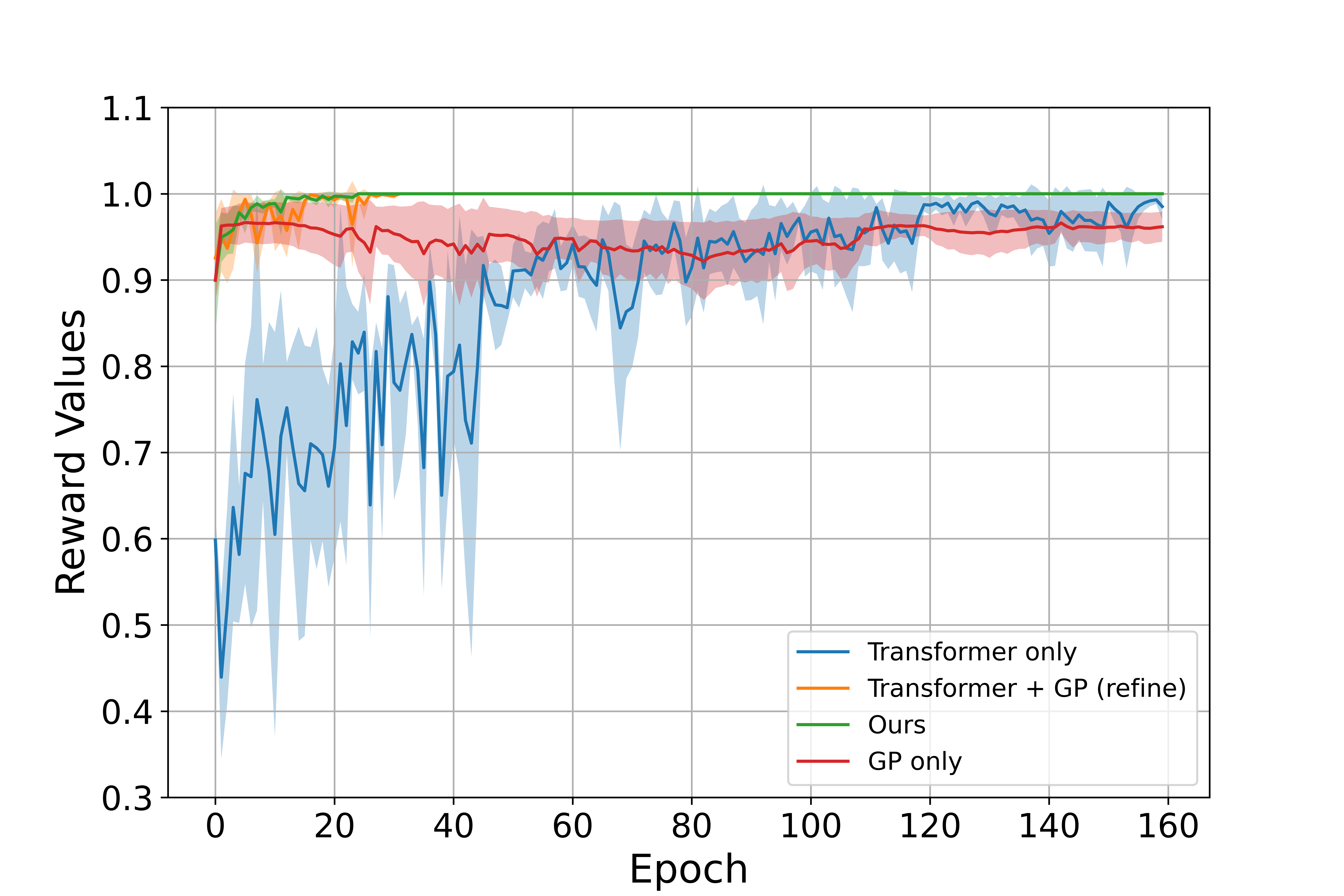}
        \caption{\textbf{Best Reward Trajectory for Ablation Study III}: Visualize the best reward trajectory of the proposed framework under four different settings in Ablation Study III on 6-D polynomial dynamics.\\}
        \label{fig:ablation-3}
    \end{minipage}
\end{figure*}

Figure \ref{fig:ablation-2} compares the highest reward trajectory in each epoch for all four settings, and Table \ref{table:ablation-verification} summarizes the training and formal verification statistics. SMT is not included in Figure \ref{fig:ablation-2} as the training stuck on the SMT solver for hours in the middle of the training stage. The proposed SHGO-based counterexample feedback makes the training reward an adversarial reward, as it can identify the most critical violations. As a `challenger', the global optimization method finds the adversarial counterexamples, not just mild violations. Consequently, this leads to a faster refinement of the policy to mitigate the critical violations and results in stronger guarantees on the final Lyapunov candidates. However, it risks training stability and may over-focus on `hard' violations.
In this case, it identifies a valid Lyapunov function $V_{1}(x) = x_{1}^{2} + x_{2}^{2} + x_{3}^{2} + x_{4}^{2} + x_{5}^{2} + x_{6}^{2}$ which is well-structured and easy to verify by formal verification tools. Note that, unlike in the experiment section, the GP component is not used here, resulting in a less structured local Lyapunov function with a few squared interaction terms $(x_i+x_j)^2$ under one random seed. This increases verification time compared to Table \ref{tab:performance summary}, yet remains significantly faster than verifying the final result from the random sampling scheme. 

Among the four settings, random sampling achieves the fastest convergence for this polynomial system, as it yields smoother rewards, covers the whole state space uniformly, and promotes training stability. However, because random sampling cannot actively search for violations—and its search space grows exponentially with dimensionality—it becomes less effective for larger or more complex systems (e.g., power systems). Even with a reward near 1, it may fail to find a valid Lyapunov function as the high reward only means it fails to locate counterexamples instead of nonexistence; or if it does, it can converge to a local minimum with poor generalization or verification properties. For instance, the discovered Lyapunov function $V_{2}(x) = x_5^2 - \cos(x_2)\cos(x_3) - \cos(x_4) - \cos(x_6) - \cos(x_1 + x_4) + 4$ has a complex shape that formal verification tools struggle to handle, and it offers limited insight beyond the training region $\mathcal{D}$. By contrast, $V_1(x)$ remains valid over the much larger domain $\mathcal{D}' = \{ (x_1, x_2, x_3, x_4, x_5, x_6) \in \mathbb{R}^6 \bm{\mid} \sum_{i=1}^{6} x_i^2 \leq 500\}$. 

Root-finding usually identifies counterexamples near function roots, producing mostly mild violations. Although an ideal root-finding algorithm might be efficient, off-the-shelf methods become inaccurate for dimensions above five. As a result, they fail to provide precise violation signals or strong optimization guidance to the transformer, causing the best reward to stall at $0.7$. Similarly, \textit{dreal} SMT often locates only mild violations rather than critical ones, limiting its impact on optimization. While these counterexamples help the transformer reduce violations, the refined transformer increasingly acts as an adversary to the SMT solver, prolonging verification until timeout. In neural network + formal verification settings, this effect is even exacerbated by the model’s overparameterization. Consequently, we rely on global optimization instead of formal verification in the training loop.

\subsection{Ablation Study III - Generic Programming and Expert Guidance Loss}
\label{app: ablation_gp}

In the proposed framework, the Genetic Programming (GP) component refines sampled candidates $\Tilde{V}_{\phi}$ from the symbolic transformer and optimizes the transformer parameters $\phi$ using the expert guidance loss in Equation \eqref{eq: weighted ce}. To assess the impact of the GP component on the training paradigm, we evaluate the following four settings of the proposed framework on the 6-D polynomial dynamics (Appendix \ref{subsec: 6d-poly}): 

\begin{itemize}
    \item \textbf{Transformer only}: A pure symbolic transformer model without GP component.

    \item \textbf{Transformer + GP}: A symbolic transformer model with a GP component for candidate refinement but without expert-guided learning.

    \item \textbf{Transformer + GP + expert-guided learning (ours)}: A symbolic transformer model with a GP component for both candidate refinement and expert-guided learning for transformer optimization.

    \item \textbf{GP only}: GP component only. The initial population of the first epoch comes from a randomly initialized symbolic transformer, and the initial population of other epochs comes from the latest refined expressions.
\end{itemize}

\begin{table*}[t!]
\centering
\caption{Number of epochs, training time, and success rate comparison between four settings of the proposed framework. Runtime is the average training time for successful trials. Succ. \% is the successful rate of finding a valid Lyapunov function out of 3 random seeds. }
\vskip 0.15in
\begin{tabular}{c|c|c|c|c}
\toprule
\multicolumn{1}{c|}{\textbf{Training Stats.}} & \multicolumn{1}{c|}{\textbf{Transformer only}} & \multicolumn{1}{c|}{\textbf{Transformer + GP}} & \multicolumn{1}{c|}{\textbf{Ours}} &
\multicolumn{1}{c}{\textbf{GP only}} \\
\midrule
\textbf{Epoch} & 168 & 20 & \textbf{18} & - \\
\textbf{Runtime} & 8399s & 2731s & \textbf{2467s} & -\\
\textbf{Succ. \%} & \textbf{100} & \textbf{100} & \textbf{100} & 0 \\
\bottomrule
\end{tabular}
\label{table:ablation-GP}
\end{table*}

\begin{figure*}
    \centering
    \begin{minipage}[t!]{0.48\textwidth}
        \centering
        \includegraphics[width=\linewidth]{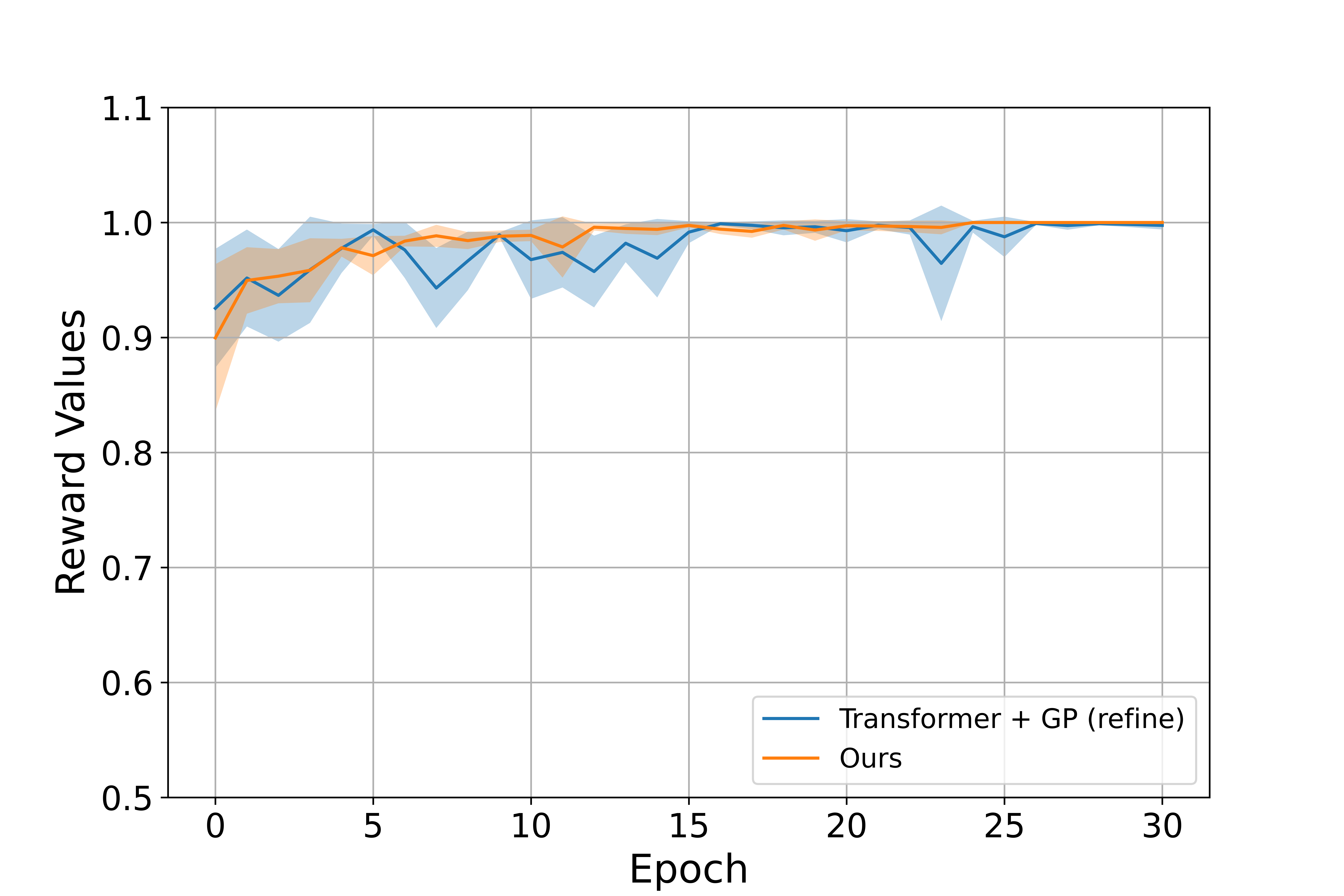}
    \end{minipage}
    \hfill
    \begin{minipage}[th!]{0.48\textwidth}
        \centering
        \includegraphics[width=\linewidth]{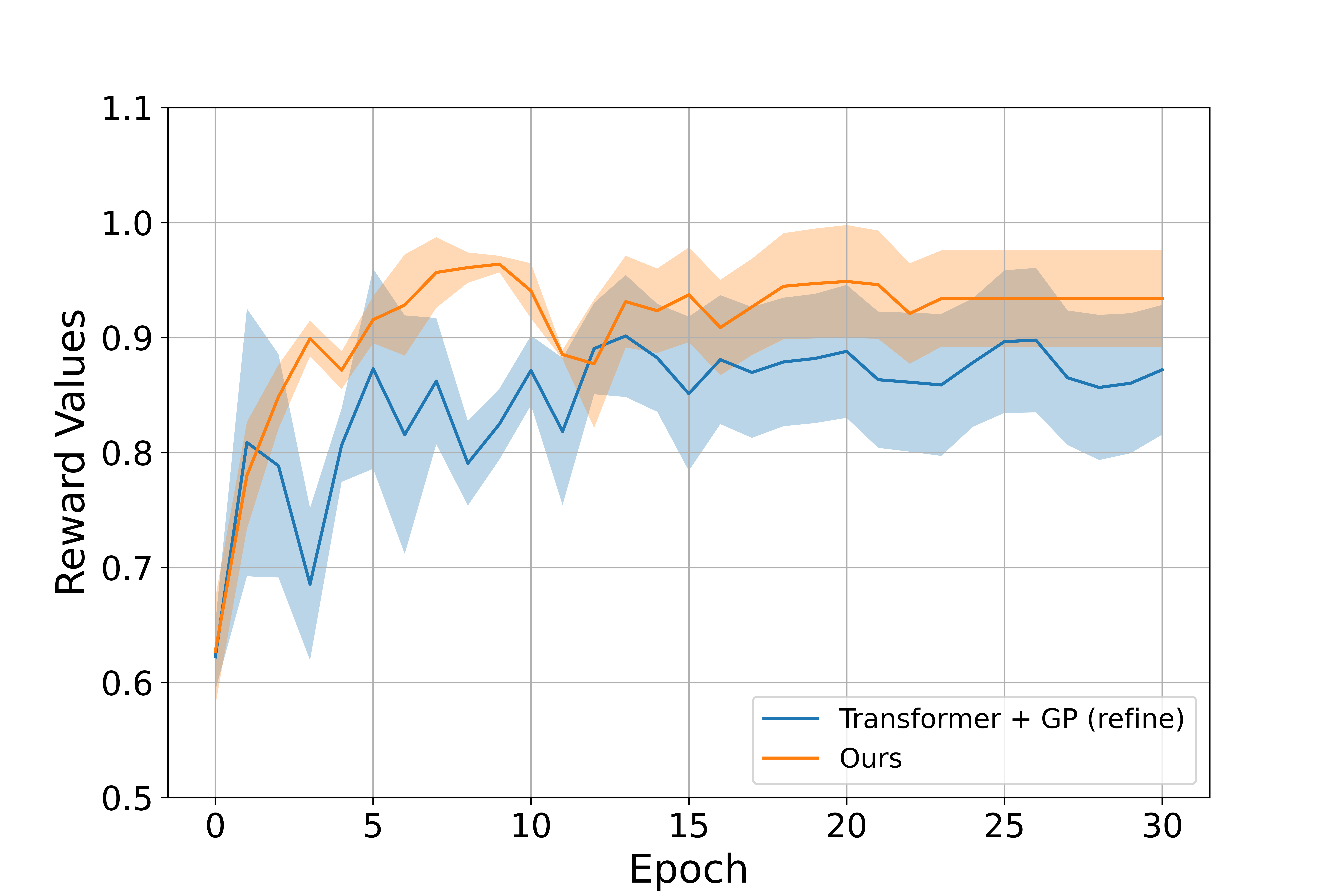}
    \end{minipage}
    \caption{The figure on the left visualizes the overall best reward (best candidate among $\Tilde{\mathcal{V}}_{\phi} \cup \Tilde{\mathcal{V}}_{gp}$) trajectory. The figure on the left visualizes the best reward trajectory for transformer candidates (best candidate among $\Tilde{\mathcal{V}}_{\phi}$ only, excludes GP-generated solutions). These two plots compare the performance of transformer + GP refinement and our method in Ablation Study III.}
    \label{fig:ablation_3_2}
\end{figure*}

Figure \ref{fig:ablation-3} shows the highest reward in each epoch for the four settings, and Table \ref{table:ablation-GP} compares their training statistics. The transformer-only approach achieves a 100\% success rate across three random seeds, proving its capability to generate Lyapunov functions on high-dimensional systems. However, without the GP component, it requires significantly more training time and epochs to identify a valid Lyapunov function because it must independently explore all necessary characteristics of a valid solution. Conversely, the GP-only approach fails to find any valid Lyapunov functions when starting from a random population, as it lacks the transformer's deeper understanding of system dynamics and relies on less capable evolutionary operations for exploration.

Figure \ref{fig:ablation_3_2} explicitly compares the reward trajectories with and without expert-guided supervised learning, where ours (with expert-guided learning) achieves the fastest convergence. Comparing the reward trajectory, ours achieves faster and more stable training. Overall, these visualizations and comparisons emphasize the benefits of GP refinement and the expert-guided learning on `elite set' $\Tilde{\mathcal{V}}_{gp}$ to accelerate transformer parameters' update and enable faster Lyapunov function discovery.


\end{document}